\documentclass{article}

\usepackage{microtype}
\usepackage{booktabs} 
\usepackage{slashbox}

\usepackage{amsmath}
\usepackage{amsfonts}
\usepackage{multirow}
\usepackage{subcaption}
\usepackage{makecell}
\usepackage{graphbox}
\usepackage{xcolor}

\usepackage{amsmath,amsfonts,bm}
\usepackage{bbm}

\def\eqref#1{equation~\ref{#1}}

\def\1{\bm{1}}
\newcommand{\train}{\mathcal{D}}

\def\rveps{\boldsymbol{\varepsilon}}

\def\rvh{{\mathbf{h}}}
\def\rvu{{\mathbf{i}}}

\def\rvm{{\mathbf{m}}}

\def\rvs{{\mathbf{s}}}

\def\rvu{{\mathbf{u}}}
\def\rvv{{\mathbf{v}}}

\def\rvx{{\mathbf{x}}}

\def\mJ{{\bm{J}}}

\DeclareMathAlphabet{\mathsfit}{\encodingdefault}{\sfdefault}{m}{sl}
\SetMathAlphabet{\mathsfit}{bold}{\encodingdefault}{\sfdefault}{bx}{n}

\def\gB{{\mathcal{B}}}

\def\gN{{\mathcal{N}}}

\def\gX{{\mathcal{X}}}

\def\sH{{\mathbb{H}}}

\newcommand{\Pdata}{P_{\rm{data}}}

\newcommand{\Ptrain}{\hat{P}_{\rm{data}}}

\newcommand{\Pmodel}{P_{\rm{model}}}

\newcommand{\indicator}{\mathbbm{1}}

\newcommand{\E}{\mathbb{E}}

\newcommand{\R}{\mathbb{R}}

\newcommand\ie{\textit{i.e.}}
\newcommand\eg{\textit{e.g.}}

\usepackage{hyperref}

\usepackage[accepted]{icml2020}

\icmltitlerunning{Learning Discrete Distributions by Dequantization}

\begin{document}

\twocolumn[
\icmltitle{Learning Discrete Distributions by Dequantization}



\icmlsetsymbol{intern}{*}

\begin{icmlauthorlist}
\icmlauthor{Emiel Hoogeboom}{uva,intern}
\icmlauthor{Taco S. Cohen}{qual}
\icmlauthor{Jakub M. Tomczak}{qual}
\end{icmlauthorlist}

\icmlaffiliation{uva}{University of Amsterdam, Netherlands}
\icmlaffiliation{qual}{Qualcomm AI Research, Qualcomm Technologies Netherlands B.V.. Qualcomm AI Research is an initiative of Qualcomm Technologies, Inc.}

\icmlcorrespondingauthor{Emiel Hoogeboom}{e.hoogeboom@uva.nl}

\icmlkeywords{dequantization,discrete distribution,density modelling,generative modelling}

\vskip 0.3in
]



\printAffiliationsAndNotice{\textsuperscript{*}Research done while completing an internship at Qualcomm AI Research, Qualcomm Technologies Netherlands. Currently a Ph.D. student at the University of Amsterdam, Netherlands.}

\begin{abstract}
Media is generally stored digitally and is therefore \emph{discrete}. Many successful deep distribution models in deep learning learn a density, \emph{i.e.}, the distribution of a \emph{continuous} random variable. Na\"{i}ve optimization on discrete data leads to arbitrarily high likelihoods, and instead, it has become standard practice to add noise to datapoints. In this paper, we present a general framework for dequantization that captures existing methods as a special case. We derive two new dequantization objectives: importance-weighted (\textit{iw}) dequantization and R\'{e}nyi dequantization. In addition, we introduce autoregressive dequantization (ARD) for more flexible dequantization distributions. Empirically we find that \textit{iw} and R\'{e}nyi dequantization considerably improve performance for uniform dequantization distributions. ARD achieves a negative log-likelihood of 3.06 bits per dimension on CIFAR10, which to the best of our knowledge is state-of-the-art among distribution models that do not require autoregressive inverses for sampling. 
\end{abstract}

\section{Introduction}
\label{sec:introduction}

Today, virtually all media is handled digitally. As such, it is stored in bits and is therefore \emph{discrete}. Deep distributions models \cite{larochelle2011nade,kingma2013autoencoding} aim to learn a distribution model $p_{\mathrm{model}}(x)$ for high-dimensional data. Many of these models are \emph{density} models \cite{uria2013rnade,oord2014factoring,dinh2017density,papamakarios2017masked}, meaning they learn a distribution of a \emph{continuous} random variable.

\begin{figure*}
\begin{subfigure}{.5\textwidth}
 \centering
  \includegraphics[width=.7\textwidth]{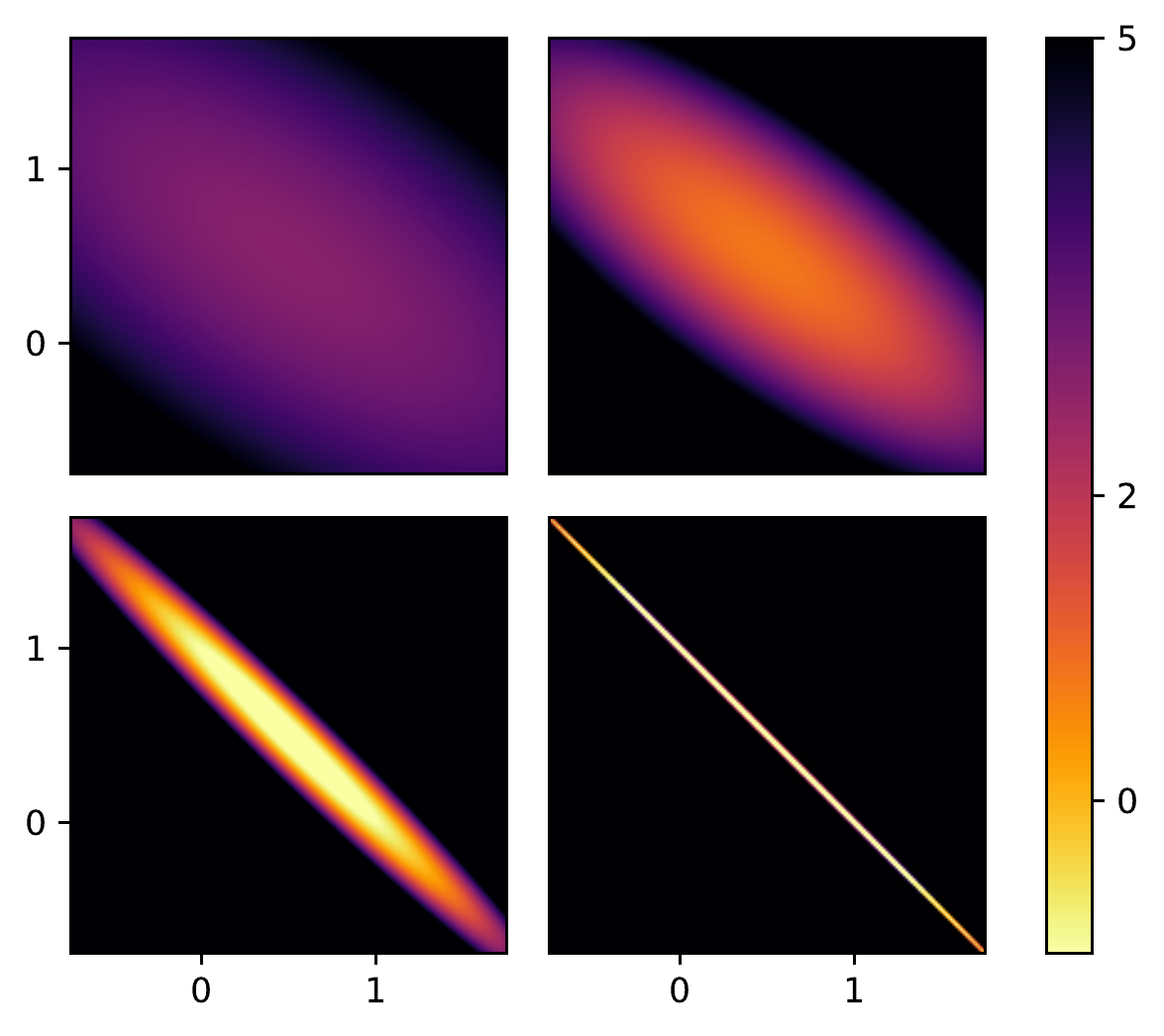}
  \caption{}
  \label{fig:naive_ll}
\end{subfigure}%
\begin{subfigure}{.5\textwidth}
  \centering
    \includegraphics[width=1.\textwidth]{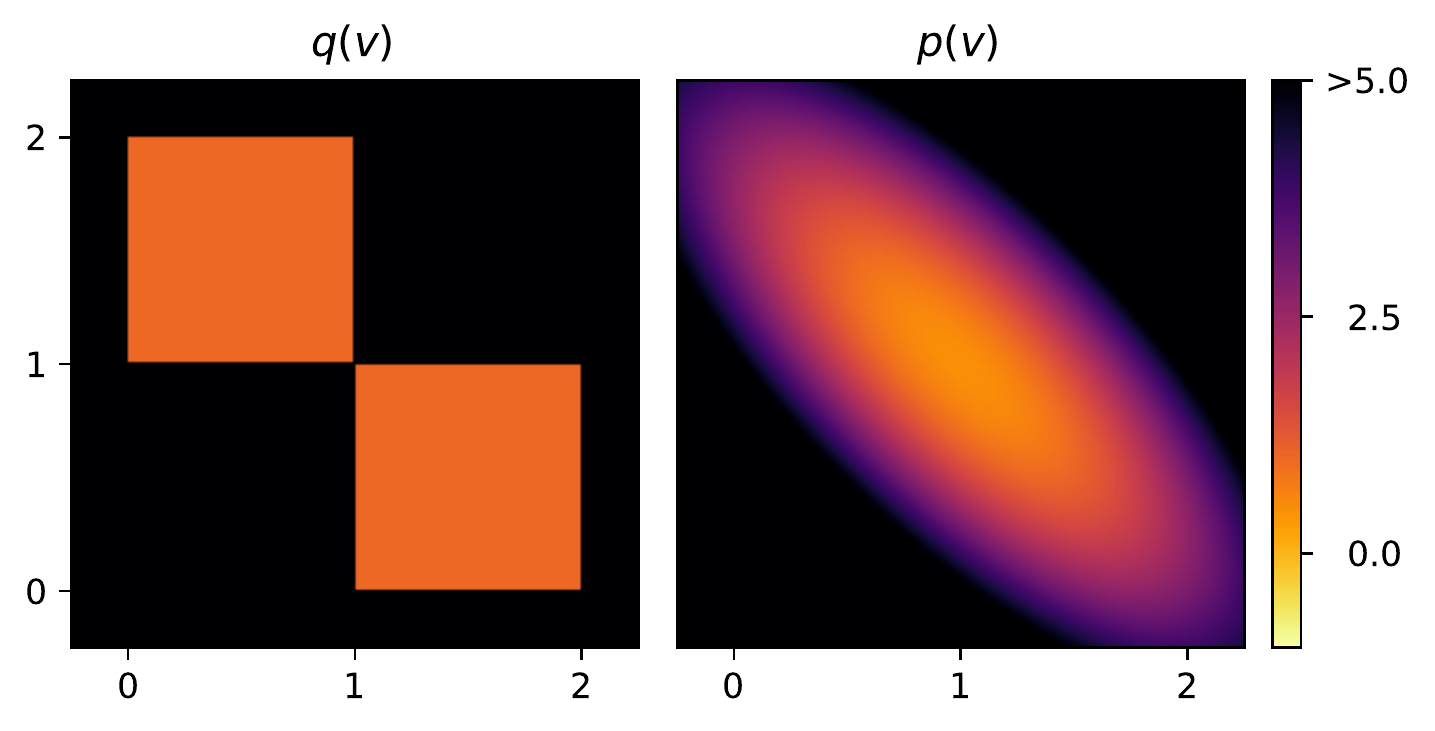}
    \caption{}
    \label{fig:naive_dequantized_ll}
\end{subfigure}
\caption{\textbf{(a)} When a continuous density model is trained on discrete data using maximum likelihood, the solution may achieve arbitrarily high likelihoods. This example, a multivariate Gaussian with full covariance matrix is fitted with gradient descent to a discrete distribution on points (0, 1) and (1, 0). The images show the model during training at iterations 20, 40, 80 and 140. Density is depicted in bits (negative log$_2$-likelihood). \textbf{(b)} The observed discrete variable can be connected to a latent continuous counterpart. In this example, simple uniform noise is added to the discrete distribution from Figure \ref{fig:naive_ll}, which \emph{dequantizes} the data (left). The multivariate Gaussian fitted to this distribution is shown on the right. Probability density is depicted in bits (negative log$_2$-likelihood).}
\label{fig:fig}
\end{figure*}

Problematically, the na\"{i}ve maximum likelihood solution for a continuous density model on discrete data, may place arbitrarily high likelihood on the discrete locations \cite{theis2016anote} (for an example see Figure \ref{fig:naive_ll}). Since discrete and continuous spaces are topologically different, a probability density does not necessarily approximate a probability mass. After all, the total probability at a single point under a density is always zero.

\begin{figure}[t]
    \centering
    \begin{subfigure}[b]{0.157\textwidth}
        \includegraphics[width=\textwidth]{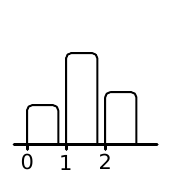}
        \caption{$P_{\mathrm{data}}(\rvx)$}
    \end{subfigure}
    \begin{subfigure}[b]{0.157\textwidth}
        \includegraphics[width=\textwidth]{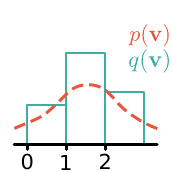}
        \caption{uniform $q(\rvv|\rvx)$}
    \end{subfigure}
    \begin{subfigure}[b]{0.157\textwidth}
        \includegraphics[width=\textwidth]{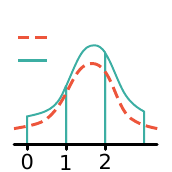}
        \caption{flexible $q(\rvv|\rvx)$}
    \end{subfigure}
    \caption{A discrete distribution $P_{\mathrm{data}}(\rvx)$ is dequantized by $q(\rvv|\rvx)$, which is visualized in the marginal continuous distribution $q(\rvv) = \mathbb{E}_{\rvx \sim P_{\mathrm{data}}}[q(\rvv|\rvx)]$. In this example the continuous density model $p(\rvv)$ is relatively simple, and two dequantization distributions $q(\rvv|\rvx)$ are considered: one is simple and the other is flexible. Suppose that the dequantization distribution $q(\rvv|\rvx)$ is uniform. Then $p(\rvv)$ is encouraged to have relatively high uncertainty under variational inference. In contrast, when the dequantization distribution $q(\rvv|\rvx)$ is flexible it can match $p(\rvv)$ which considerably improves the tightness of the variational bound.}
    \label{fig:overview}
\end{figure}



To deal with this issue, it has become common practice to add noise to datapoints which \textit{dequantizes} the data. \citet{theis2016anote} show that if noise is added in a particular way, the likelihood from the continuous model is a lowerbound of the discrete model (for an example see Figure \ref{fig:naive_dequantized_ll}). This is important as it allows comparison of discrete and continuous models directly using the \textit{likelihood}. Recently \citet{ho2019flowpp} show that improving the flexibility of the noise distribution allows tighter bounds which improves modelling performance.

Although the benefits of learned dequantization have been demonstrated in a specific case, the effects of dequantization are not yet fully understood. How do dequantization and density model interact? What is the effect of increased dequantization flexibility? Are there more sophisticated optimization objectives? 

In this paper, we present a general framework for dequantization via latent variable modelling. In this framework, we are able to recover existing dequantization schemes as special cases, and we derive two new objectives: importance-weighted dequantization, and R\'{e}nyi dequantization. In addition, we propose autoregressive flows to learn dequantization distributions. Although autoregressive flows are computationally expensive to invert, in this particular case the dequantization noise does not have to be inverted. Experimentally we show how density and dequantization distributions with varying flexibility interact on a 2-dimensional problem. In addition, we find that our methods improve likelihood modelling on binary MNIST and CIFAR10.

The contribution of the paper is threefold: 
\begin{itemize}
    \item We outline a latent variable framework for latent variable dequantization. We recover variational inference (\textit{vi}) dequantization \citep{ho2019flowpp} and propose two new dequantization approaches based on the weighted importance sampling (\emph{iw} dequantization) and variational R\'{e}nyi divergence (R\'{e}nyi dequantization).
    \item We outline different dequantization distributions. We opt for using autoregressive flow dequantization (ARD) which consistently improves variational inference and log-likelihood evaluation of density models. Even though ARD utilizes autoregressive modules, it is possible to sample from the model without computing the inverse of autoregressive modules.
    \item We evaluate \textit{iw}, R\'{e}nyi  and \textit{vi} dequantization and test different dequantization distributions on image datasets (binary MNIST and CIFAR10) quantitatively. Furthermore, we analyze the learned densities for a 2d data problem qualitatively. Using experimental results, we describe recommendations for which dequantization methods to utilize. 
\end{itemize}

\section{Related Work}
\label{sec:related_work}

Often, probability distributions of continuous variables are parameterized by neural networks \citep{kingma2013autoencoding, rezende2014stochastic, dinh2017density}. Many distributions are discrete, \textit{e.g.}, in 8 bit images the pixels take values in \{0, 1, \dots, 255\}. A probability distribution over discrete variables can be modeled using continuous latent-variable models like Variational Auto-Encoders (VAEs) \citep{kingma2013autoencoding, rezende2014stochastic} or directly by applying autoregressive models (ARMs) \citep{oord2016pixel}. VAEs are rather easy to train and could be parameterized using different neural network architectures, however, they provide a lowerbound to the log-likelihood. On the other hand, ARMs provide an exact value of the marginal likelihood in a fast manner, but they are typically slow to sample from.

Further, flow-based models have recently also been applied to discrete variable modelling \cite{tran2019discrete,hoogeboom2019integer}. \citet{tran2019discrete} consider binary and categorical variables in text analysis, but their performance on image data is currently unknown. \citet{hoogeboom2019integer} show competitive performance on image data.

However, a large number of distribution models learn a density, a distribution over a continuous variable \cite{uria2013rnade,oord2014factoring,dinh2017density,papamakarios2017masked,kingma2018glow,huang2018neural,cao2019block,grathwohl2019ffjord,hoogeboom2019emerging,ho2019flowpp,chen2019residual,song2019mintnet,ma2019macow}. A standard approach adds uniform noise to discrete values \cite{theis2016anote, uria2013rnade, oord2014factoring}. Very recently, it was proposed to consider a learnable dequantization treated as a variational posterior over latent variables (\textit{i.e.}, continuous variables) \citep{ho2019flowpp, winkler2019learning}. In this paper, we derive a new framework for dequantization using latent variable modelling and we present two new dequantization objectives. We provide more in-depth analysis and aim at understanding how different choices of dequantization objectives and dequantization distributions affect the final performance in the log-likelihood.

\section{Methodology}
\label{sec:method}

Let $\rvx \in \gX$ denote a vector of $D$ observable discrete random variables and $\Pdata(\rvx)$ be its (unknown) distribution. We assume there is a set of data $\train=\{\rvx_{n}\}$ given, or, equivalently, an empirical distribution $\Ptrain (\rvx)$ is provided.
The likelihood-based approach to learning a distribution is about finding values of parameters of a model $\Pmodel(\rvx)$ that maximize the log-likelihood function:
\begin{equation}
    \log \Pmodel(\train) = \E_{\rvx \sim \Ptrain(\rvx)}[\log \Pmodel(\rvx_n)].
\end{equation}

\subsection{Dequantization as a latent variable model}
Frequently, a discrete distribution models a proxy of a continuous variable in the physical world. For instance, a digital photograph of an observed scene represents the light that is reflected from observed objects, quantized to a certain precision.
In other words, we can consider a latent variable model where continuous latent variables $\rvv \in \R^{D}$ correspond to a continuous representation of the world and observable discrete variables $\rvx$ are measured quantities.
This suggests the following model:
\begin{equation}\label{eq:quantizer_continuous}
    \Pmodel(\rvx) = \int P_{\vartheta}(\rvx|\rvv) p_{\theta}(\rvv) \mathrm{d} \rvv,
\end{equation}
where $P_{\vartheta}(\rvx|\rvv)$ is an indicator function of $\rvv$ being contained in a volume $\gB_{\vartheta}(\rvx) \subseteq \R^{D}$, namely, $P_{\vartheta}(\rvx|\rvv) = \indicator[\rvv \in \gB_{\vartheta}(\rvx)]$, and $p_{\theta}(\rvv)$ is a continuous distribution, which may be modeled using a flexible density model \cite{mackay1999density, dinh2017density, rippel2013high}.
We refer to $P_{\vartheta}(\rvx|\rvv)$ as a \textit{quantizer}. Note that in principle the volumes $\gB_{\vartheta}$ can be constructed to induce any type of partition of a volume space, where care should be taken that $\gB_{\vartheta}$ for different $\rvx$ do not overlap. When we set $\gB(\rvx) = \{\rvx \cdot \rvu | \rvu \in \mathbb{R}_{+}^{D}\}$ for $\rvx \in \{-1, 1\}^D$ we recover half-infinite dequantization for binary variables from \citet{winkler2019learning}. In this paper, since image data is often represented on a square grid we will focus on \emph{hypercubes}, namely, $\gB(\rvx) = \{\rvx + \rvu : \rvu \in [0, 1)^{D}\}$.

Calculating the integral in (\ref{eq:quantizer_continuous}) is troublesome, and thus, learning is infeasible especially in high dimensional cases.
Therefore, in order to alleviate this issue, we introduce a new distribution $q_{\phi}(\rvv|\rvx)$ with parameters $\phi$, a \textit{dequantizing} distribution or \textit{dequantizer}.
In fact, the dequantizer should have the same support as $P_{\vartheta}(\rvx|\rvv)$, otherwise it would assign probability mass to regions outside the volume $\gB(\rvx)$.
Therefore, we will use $\rvu$ instead of $\rvv$ in the dequantizing distribution to highlight the fact that the support of $q_{\phi}(\rvv|\rvx)$ equals $\gB(\rvx)$, where we define $\rvv = \rvx + \rvu$.
Including the dequantizer in our model yields:
\begin{align}
    \Pmodel(\rvx) &= \int \frac{q_{\phi}(\rvu|\rvx) P_{\vartheta}(\rvx|\rvv) p_{\theta}(\rvv)}{q_{\phi}(\rvu|\rvx)} \mathrm{d} \rvv \\
                  &= \E_{\rvu \sim q_{\phi}(\rvu|\rvx)} \Big{[} \frac{P_{\vartheta}(\rvx|\rvv) p_{\theta}(\rvv)}{q_{\phi}(\rvu|\rvx)} \Big{]} \label{eq:dequantization},
\end{align}
Existing methods in literature define $P(\rvx) = \int_{[0, 1)^D} p(\rvx + \rvu) \mathrm{d}\rvu$ to relate a discrete and continuous model. Important differences with our method is that ours can be derived directly without this definition, and the quantizer volume $\gB$ generalizes to any volumetric partition.

Introducing the dequantizer allows us to derive three approaches to approximate the integral using \emph{i)} variational inference, \emph{ii)} weighted importance sampling and \emph{iii)} variational R\'{e}nyi approximation.
    
\subsection{Variational Dequantization}

We can interpret the dequantizing distribution as a \textit{variational} distribution and apply Jensen's inequality to obtain the lower-bound on the log-likelihood function:
\begin{equation}\label{eq:elbo}
    \log \Pmodel(\rvx) \geq \E_{\rvu \sim q_{\phi}(\rvu|\rvx)} \Big{[} \log  \frac{P_{\vartheta}(\rvx|\rvv) p_{\theta}(\rvv)}{q_{\phi}(\rvu|\rvx)} \Big{]} .
\end{equation}
The dequantizing distribution must be restricted to assign probability mass to $\gB(\rvx)$ only, otherwise the lower-bound is undefined.
Thus, for our choice of $\gB(\rvx)$ being a hypercube, we can apply the sigmoid function to the output of the dequantizer to ensure the lower-bound has appropriate support.
As a result, we can re-write (\ref{eq:elbo}) as follows:
\begin{equation}\label{eq:variational_dequantization}
    \log \Pmodel(\rvx) \geq \E_{\rvu \sim q_{\phi}(\rvu|\rvx)} \Big{[} \log  p_{\theta}(\rvv) \Big{]} + \sH [q_{\phi}],
\end{equation}
which recovers the variational dequantization from \citet{ho2019flowpp}. Note that $\sH [q_{\phi}] = -\E_{\rvu \sim q_{\phi}(\rvu|\rvx)} \Big{[} \log  q_{\phi}(\rvu|\rvx) \Big{]}$ is the entropy of the dequantizing distribution for a given $\rvx$, which prevents the dequantizer from collapsing to a delta peak.  We refer to this dequantization scheme as \emph{vi dequantization}.

\subsection{Importance-Weighted Dequantization}
Alternatively, we can interpret the dequantizing distribution as a \textit{proposal} distribution and instead of using Jensen's inequality we sample $K$ times from $q_{\phi}(\rvu|\rvx)$, which directly approximates the log-likelihood:
\begin{equation}\label{eq:iw_general}
    \log \Pmodel(\rvx) \geq \log \Big{[} \frac{1}{K} \sum_{k=1}^{K} \frac{P_{\vartheta}(\rvx|\rvv_{k}) p_{\theta}(\rvv_{k})}{q_{\phi}(\rvu_{k}|\rvx)} \Big{]} ,
\end{equation}
where $\rvu_{k} \sim q_{\phi}(\rvu|\rvx)$ and $\rvv_k = \rvx + \rvu_{k}$ for $k=1, 2, \ldots, K$.
If we constrain the proposal distribution (the dequantizer) in the same manner as we did in the case of the variational dequantization (\ie, the probability mass should be assigned only to $\gB(\rvx)$), we obtain: 
\begin{align}
    \log \Pmodel(\rvx) &\geq \log \Big{[} \frac{1}{K} \sum_{k=1}^{K} \frac{p_{\theta}(\rvv_{k})}{q_{\phi}(\rvu_{k}|\rvx)} \Big{]} \label{eq:iw_specific}\\
        &= \log \Big{[} \frac{1}{K} \sum_{k=1}^{K} w_{k}(\rvx) \Big{]},
\end{align}
where $w_{k}(\rvx) \stackrel{\Delta}{=} \frac{p_{\theta}(\rvv_{k})}{q_{\phi}(\rvu_{k}|\rvx)}$ is an importance weight.

In general, if $K \rightarrow \infty$, then we obtain an equality in (\ref{eq:iw_general}) and (\ref{eq:iw_specific}).
But since we take a finite sample, the approximate gives a lower-bound to the log-likelihood function (\textit{iw-bound}).
Importantly, the iw-bound is tighter than the variational lower-bound \cite{burda2015importance, domke2018importance}.
Hence, the importance-weighting is preferable over the variational inference and in practice it leads to a better log-likelihood performance. We refer to this dequantization scheme as \emph{iw dequantization}.

\subsection{R\'{e}nyi Dequantization}
The variational inference and importance-weighting sampling for a latent variable model could be generalized by noticing that both approaches are special cases of the variational R\'{e}nyi bounds.
It has been shown in \cite{li2016renyi} that the log-likelihood function could be lower-bounded by the R\'{e}nyi divergence approximated with the sample from $q_{\phi}(\rvu|\rvx)$ of size $K < \infty$, namely:
\begin{align}\label{eq:vr_general}
    \log& \Pmodel(\rvx) \geq \notag\\
    &\frac{1}{1-\alpha}\log \Big{[} \frac{1}{K} \sum_{k=1}^{K} \Big{(} \frac{P_{\vartheta}(\rvx|\rvv_{k}) p_{\theta}(\rvv_{k})}{q_{\phi}(\rvu_{k}|\rvx)} \Big{)}^{1-\alpha} \Big{]} ,
\end{align}
where $\alpha \in [0, 1)$ is a hyperparameter.
Interestingly, for $\alpha \rightarrow 1$ we obtain the variational lower-bound and for $\alpha = 0$ we get the iw-bound.

\citet{li2016renyi} have further shown that it is advantageous to consider $\alpha < 0$, because it may give tighter bounds than the iw-bound when the sample size $K$ is low.\footnote{To be precise, if we consider the infinite sample for $\alpha < 0$, we get an upper-bound on the log-likelihood function.
However, taking $K < \infty$ results in a tight lower-bound according to Corollary 1 in \cite{li2016renyi}.}
Setting $\alpha = - \infty$ corresponds to picking the largest importance weight value.
Using the notation introduced in \ref{eq:iw_specific}, we can obtain the variational R\'{e}nyi max approximation (\textit{VR-max}):
\begin{equation}\label{eq:vr_max_bound}
    \log \Pmodel(\rvx) \approx \log \Big{[} \max_{k = 1, 2, \ldots, K} w_{k}(\rvx) \Big{]}.
\end{equation}

The maximum weight dominates the contributions of all the gradients \cite{li2016renyi}.
Therefore, the VR-max approach could be seen as a fast approximation to the importance-weighting.
The VR-max approximation speeds up computations by considering only one example instead of $K$ in calculating gradients.
We refer to this whole dequantization scheme as \textit{R\'{e}nyi dequantization}.

\subsection{Dequantizing distributions}
\label{sect:dequantizing}

The dequantizing distribution plays an important role in the framework and its flexibility allows to obtain better log-likelihood scores.
As already noticed by \cite{ho2019flowpp}, replacing a simple uniform distribution with a more sophisticated bipartite flow gives much better results.
Importantly, the dequantizing distribution is a conditional distribution and we use it for sampling instead of calculating probabilities.
Therefore, we can utilize models that are more powerful, but typically slow for evaluating probabilities, \eg, autoregressive flows \cite{kingma2016improving}. 

\paragraph{Uniform Dequantization} The special case in which $q_{\phi}(\rvu|\rvx)$ is a uniform distribution over $\gB(\rvx)$, is equivalent to the setting introduced in \cite{theis2016anote, uria2013rnade, oord2014factoring}, termed \textit{uniform dequantization}.

\paragraph{Gaussian Dequantization} A more powerful dequantization scheme than the uniform dequantization is a conditional logit-normal distribution \cite{atchison1980logistic}, namely, $q_{\phi}(\rvu|\rvx) = \mathrm{sigm} \Big{(} \gN\big{(}\mu_{\phi}(\rvx), \Sigma_{\phi}(\rvx)\big{)} \Big{)}$, where $\mu_{\phi}(\rvx)$ and $\Sigma_{\phi}(\rvx)$ denote the mean and the covariance matrix for given $\rvx$, respectively, and $\mathrm{sigm}(\cdot)$ is the sigmoid function.

\paragraph{Flow-based Dequantization}  Instead of using a certain family of distribution, we can define the quantizer by applying the change of variables formula, that is:
\begin{equation}
    q_{\phi}(\rvu|\rvx) = q_{\phi}\big{(} \rveps = f_{\phi}(\mathrm{sigm}^{-1}(\rvu);\rvx)  | \rvx\big{)} |\mJ| ,
\end{equation}
where $f: \R^{D} \rightarrow \R^{D}$ is a bijective map to a simple base distribution $q_\phi(\rveps|\rvx)$, and $\mJ = \frac{\partial \rveps}{\partial \rvu}$ denotes a Jacobian matrix. Notice we highlight the need of using the (inverse) sigmoid function on top of the bijective map in order to ensure that $\rvu \in [0, 1)^{D}$.

There are two important parts of a flow-based model, namely, a choice of a \textit{base distribution} $q_{\phi}(\rveps|\rvx)$ and a form of the bijective map $f_{\phi}$. Here we decide to use a diagonal Gaussian base distribution \cite{dinh2017density} and we present two common choices of constructing $f_{\phi}$: \emph{i)} bipartite bijective maps, and \emph{ii)} autoregressive bijective maps.

\textit{Bipartite Dequantization} The idea behind the bipartite bijective maps is to ensure invertibility by splitting an input into two parts (\eg, along channels), $\rveps = [\rveps_1, \rveps_2]$, and processing only the second part \cite{dinh2017density}, namely:
\begin{align}\label{eq:coupling_layer}
    \begin{aligned}
        \rvu_{1} &= \rveps_1 \\
        \rvu_{2} &= s_{\phi}(\rveps_1 ; \rvx) \odot \rveps_2 + t_{\phi}(\rveps_1 ; \rvx)
    \end{aligned}
\end{align}
where $s_{\phi}(\rveps_1 ; \rvx)$ is a scaling transformation, $t_{\phi}(\rveps_1 ; \rvx)$ is a translation, and $\odot$ denotes an element-wise multiplication.
We explicitly write the dependency on $\rvx$ to indicate how we use the conditioning in the dequantizer. 
The transformation in (\ref{eq:coupling_layer}) is called a \textit{coupling layer}.

Further, in order to ensure that all random variables are processed, the outputs of a coupling layer are permuted and another coupling layer is applied.

\textit{Autoregressive Dequantization} We can model $q_{\phi}(\rvu|\rvx)$ with an `expensive to invert' bijective map. 
In this paper, we find that an autoregressive model as proposed for variational autoencoders \cite{kingma2016improving} is an appealing choice for dequantization. The model could be formulated as follows:
\begin{align}\label{eq:iaf}
    \begin{aligned}
        \left[ \rvm , \rvs \right] &= \mathrm{ARM}_{\phi}(\rveps, \rvh) \\
        \rvu &= \rvs \odot \rveps + \rvm
    \end{aligned}
\end{align}
where $\mathrm{ARM}_{\phi}$ is an autoregressive model (an autoregressive neural network), $\rvh$ is a context variable that is calculated based on the conditioning $\rvx$ using a neural network, $\rvs$. We refer to this dequantization scheme as \textit{Autoregressive Dequantization (ARD)}.

\subsection{Continuous distributions}
\label{sect:continuous_distribution}

The continuous model $p_{\theta}(\rvv)$ is the crucial component in the presented framework since the better performance depends on the flexibility of this model.
In principle, any continuous density model could be used as $p_{\theta}(\rvv)$, \eg, any model mentioned in \autoref{sect:dequantizing}.
In practice, however, $p_{\theta}(\rvv)$ has to be evaluated during training \emph{and} we are interested in sampling $\rvv \sim p_{\theta}(\rvv)$. Hence, utilizing models with autoregressive components would be prohibitively slow.
Therefore, in our experiments, we consider a Gaussian distribution with diagonal covariance, full covariance, and a bipartite flow-based model (a series of coupling layers and a factored-out base distribution) as a continuous distribution.

\section{Experiments}
\label{sec:experiments}

\paragraph{Data} To understand and evaluate different dequantization schemes, they are tested on three different data problems: \emph{i)} a 2-dimensional binary problem, \emph{ii)} (statically) binarized MNIST \cite{larochelle2011nade} which is derived directly from MNIST and \emph{iii)} CIFAR10 \cite{krizhevsky2009learning} (8 bit and 5 bit). Generally we find that for problems with lower bit depths dequantization matters more for performance, as the range of dequantization noise $\rvu$ is relatively large with respect to the range of the data $\rvx$. For MNIST data we use the given split of $40000$ train, $10000$ validation and $10000$ test images. For CIFAR10 we split the 50000 training images into the first $40000$ for train and the last $10000$ for validation, we use the $10000$ test images as provided. 

\paragraph{Evaluation} Performance is evaluated on a held-out test-set using negative log-likelihood. This method of evaluation is common in deep distribution learning literature because it allows for an information theoretic interpretation: the negative log$_2$-likelihood is expressed in \emph{bits} or \emph{bits per dimension} (bpd), where the latter is an average over dimensions. Interestingly, this number represents the theoretical lossless compression limit when this model is used to compress the data.

\paragraph{Details} In the experiments we consider diagonal Gaussian, covariance Gaussian and flows as distribution models, since these models admit exact likelihood evaluation. The diagonal Gaussian is parametrized striaghtforwardly using parameters for mean and log scale. The covariance Gaussian is parametrized using a Cholesky decomposition, \textit{i.e.}, the precision $\Lambda = \mathrm{L L}^\mathrm{T}$ where $\mathrm{L}$ is the learnable parameter. The diagonal of $\mathrm{L}$ is modelled separately using a log~diagonal parameter, which ensures positive-definiteness of $\Lambda$. The covariance matrix is defined then as $\Sigma = \Lambda^{-1}$. Further, flows have an architecture as described in \cite{kingma2018glow} using the coupling networks from \cite{hoogeboom2019integer}. For more details regarding architecture and training details, see Appendix \ref{sec:architecture_and_optimization_details}.

\subsection{Analysis in 2d}
\begin{figure*}
    \centering
    \begin{subfigure}[b]{0.4\textwidth}
        \includegraphics[width=\textwidth]{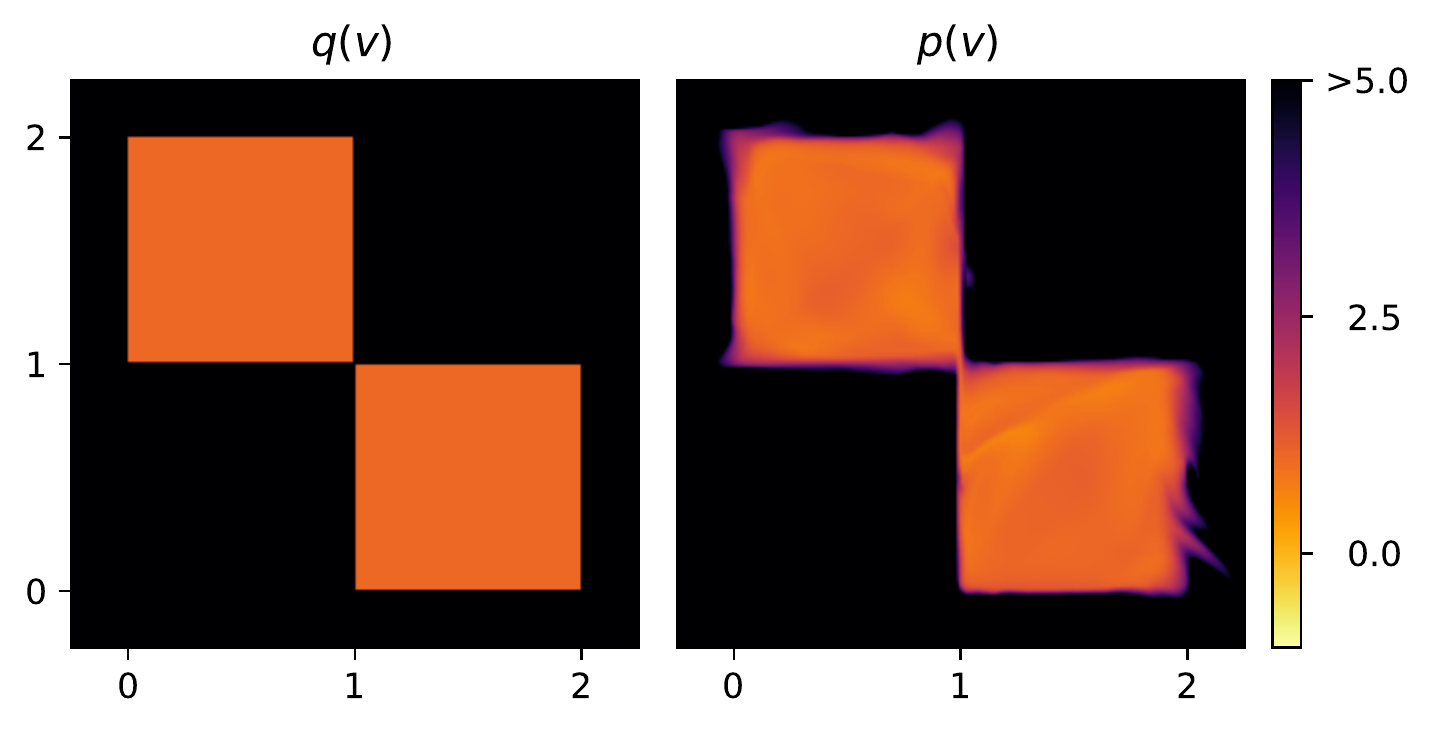}
        \caption{$q(\rvv|\rvx)$ is uniform, $p(\rvv)$ is a flow, 1.11 bits.}
        \label{fig:binary_checkerboard_density_plot_uniform}
    \end{subfigure}
    \hspace{1.cm}
    \begin{subfigure}[b]{0.4\textwidth}
        \includegraphics[width=\textwidth]{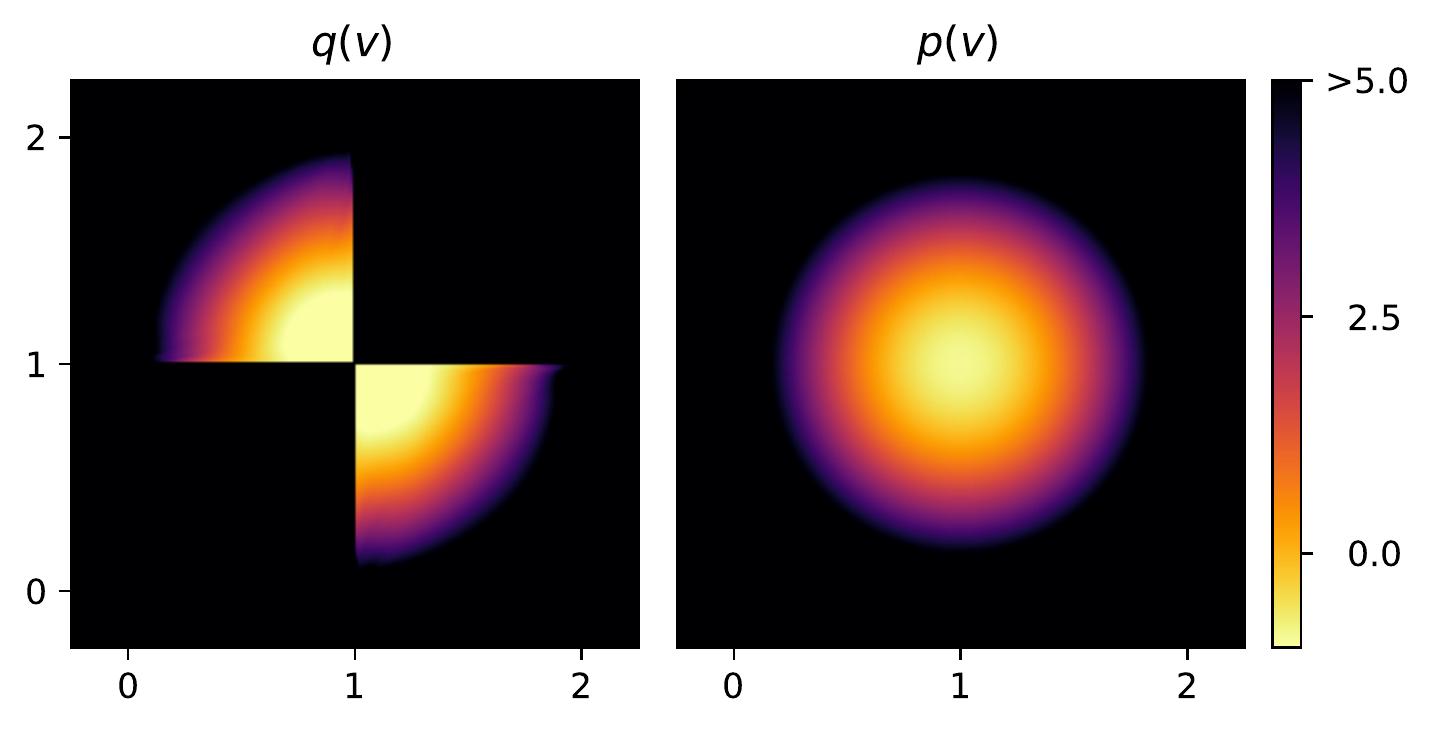}
        \caption{$q(\rvv|\rvx)$ is a flow, $p(\rvv)$ is a diag. normal, 2.08 bits.}
        \label{fig:binary_checkerboard_density_plot_diagonal}
    \end{subfigure}
    \begin{subfigure}[b]{0.4\textwidth}
        \includegraphics[width=\textwidth]{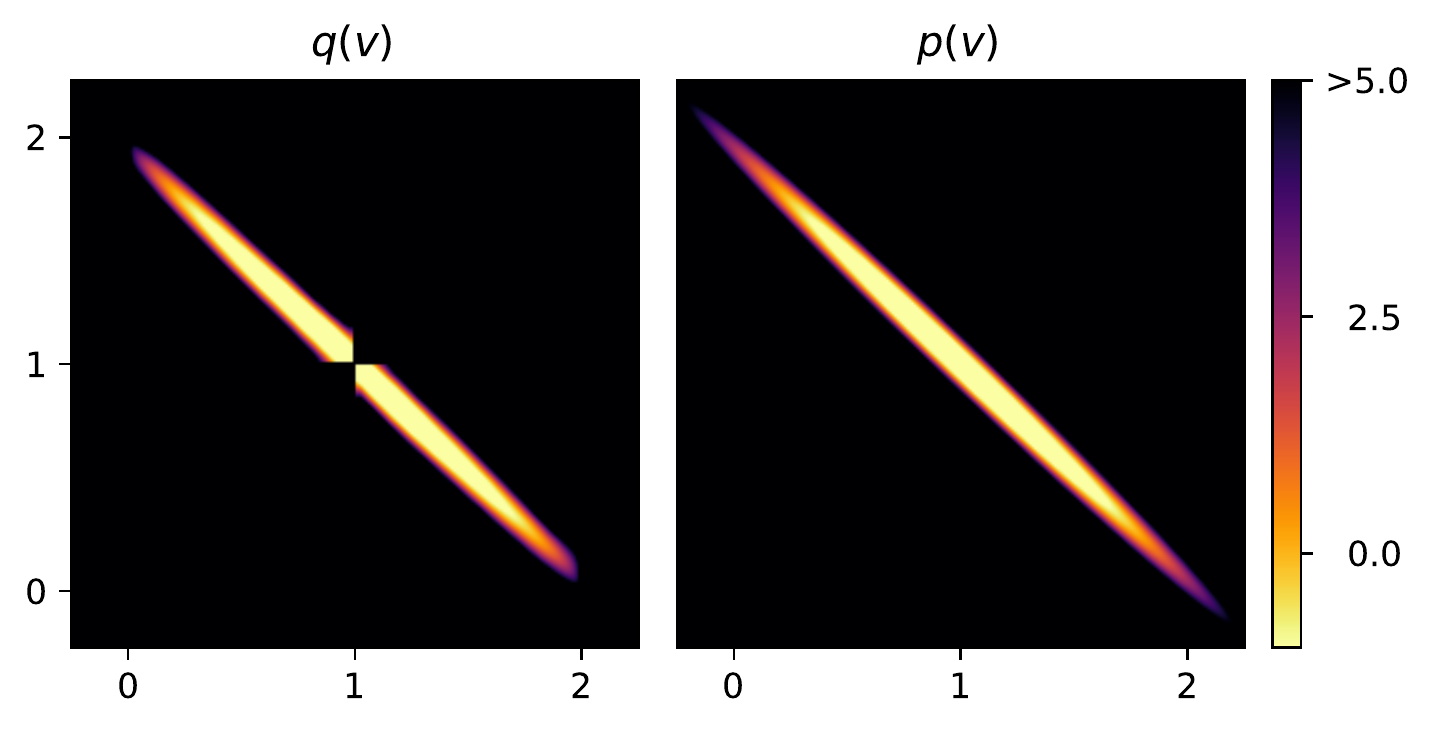}
        \caption{$q(\rvv|\rvx)$ is a flow, $p(\rvv)$ is a cov. normal, 1.08 bits.}
        \label{fig:binary_checkerboard_density_plot_cov}
    \end{subfigure}
    \hspace{1.cm}
    \begin{subfigure}[b]{0.4\textwidth}
        \includegraphics[width=\textwidth]{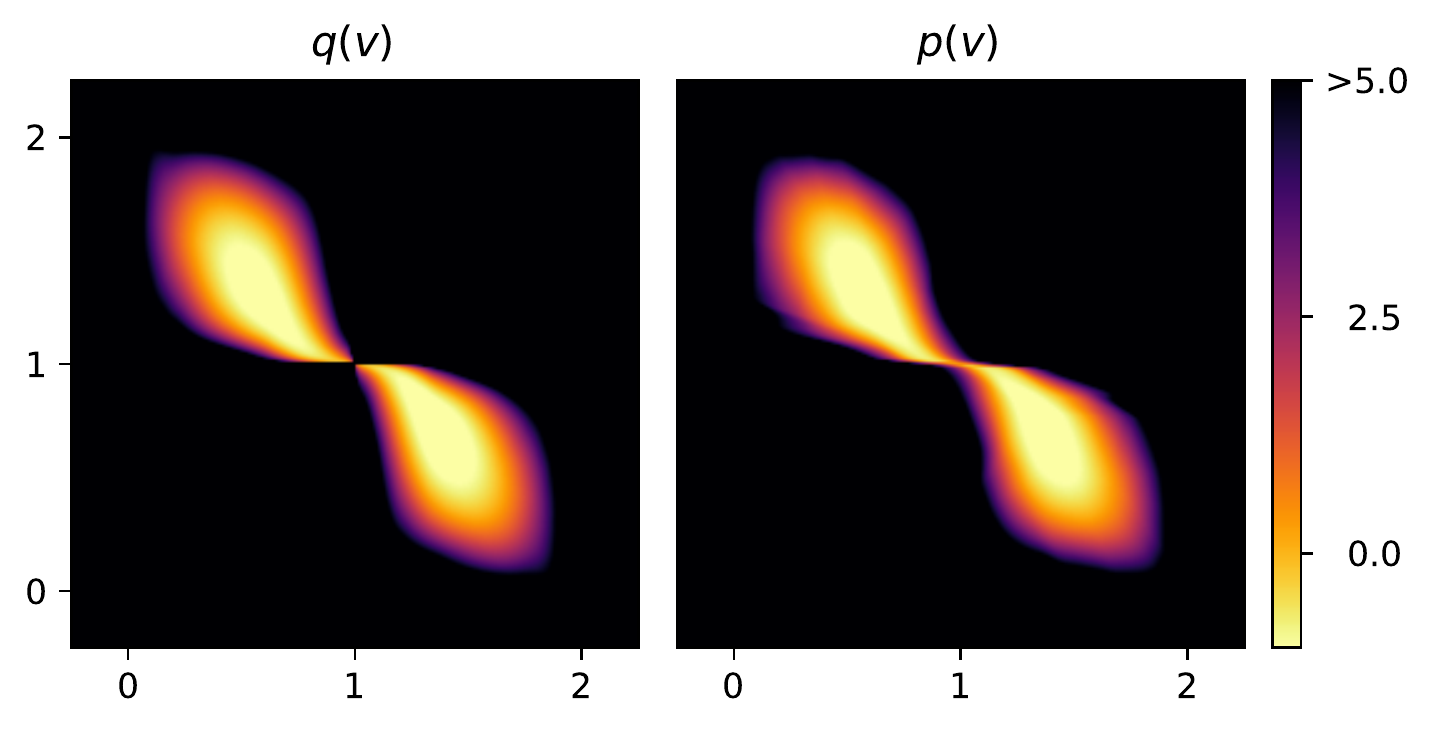}
        \caption{$q(\rvv|\rvx)$ is a flow, $p(\rvv)$ is a flow, 1.02 bits.}
        \label{fig:binary_checkerboard_density_plot_flexible}
    \end{subfigure}
    \caption{Density visualization of different density models $p(\rvv)$ and dequantizer $q(\rvv|\rvx)$ pairs. This figure considers a selection, for all different pairs we tested please see Appendix \ref{sec:visualizations}. The dequantizing distributions is visualized in the marginal distribution $q(\rvv) = \mathbb{E}_{\rvx \sim \Pdata(\rvx)} [q(\rvv|\rvx)]$. Models are trained using \textit{vi}-dequantization and the values reported are \textit{vi} evaluation.}
    \label{fig:binary_checkerboard_density_plot}
\end{figure*}

First, we analyze different dequantization methods and objectives in two dimensions. In two dimensions the learned dequantization and model distribution can be visualized. We construct a binary problem named the \emph{binary checkerboard}, which places uniform probability over two of the four states in the binary space $\{0, 1\}^2$:
\begin{equation}
\Pdata(x_1, x_2) = 
\begin{cases}
0 \quad &\text{if} \quad x_1 = 0,\, x_2 = 0, \\
\frac{1}{2} \quad &\text{if} \quad x_1 = 1,\, x_2 = 0, \\
\frac{1}{2} \quad &\text{if} \quad x_1 = 0,\, x_2 = 1, \\
0 \quad &\text{if} \quad x_1 = 1,\, x_2 = 1. \\
\end{cases}
\label{eq:binary_checkerboard_data}
\end{equation}
The theoretical likelihood limit of a dataset is typically unknown, however, the binary checkerboard is artificially constructed. Hence, the theoretical limit of the average negative log$_2$-likelihood is known and it is exactly 1 bit for this problem, because there is an equal probability over two events. In particular:
\begin{align}
\begin{split}
\mathbb{E}_{\rvx \sim P_{\textrm{data}}} &\left[ -\log \E_{\rvu \sim q_{\phi}(\rvu|\rvx)} \Big{[} \frac{p_{\theta}(\rvv)}{q_{\phi}(\rvu|\rvx)} \Big{]} \right] \\ 
&= \mathbb{E}_{\rvx \sim P_{\textrm{data}}}\Big{[}-\log P_{\mathrm{model}}(\rvx) \Big{]} \geq \sH[P_{\textrm{data}}],
\end{split}
\end{align}
where the first equality becomes an inequality when approximated with variational inference ($vi$) or a finite sample importance-weighting ($iw$). The optimum of the objective is reached when $P_{\mathrm{model}} = P_{\mathrm{data}}$, in that case the cross-entropy equals the entropy $\sH[P_{\textrm{data}}]$. For the binary checkerboard specifically, $\sH[P_{\textrm{data}}] = 1$ bit.

\paragraph{Visual analysis} Since the problem is two dimensional, the learned distributions can be visualized. Figure \ref{fig:binary_checkerboard_density_plot} depicts the probability density of the dequantizer $q(\rvv|\rvx)$ and the density model $p(\rvv)$, for models trained using \textit{vi}-dequantization. Since by construction $q(\rvv|\rvx)$ only places density on a bin corresponding to $\rvx$, the distribution $q(\rvv|\rvx)$ can be visualized without overlap in the marginal distribution $q(\rvv) = \mathbb{E}_{\rvx \sim \Pdata(\rvx)} [q(\rvv|\rvx)]$.

When the model $p(\rvv)$ is a flow and the dequantizer $q(\rvv|\rvx)$ is uniform (Figure \ref{fig:binary_checkerboard_density_plot_uniform}), the model $p(\rvv)$ is struggling to adequately model the boundaries of the dequantized density regions. When the model $p(\rvv)$ is a simple diagonal Gaussian and the dequantizer $q(\rvv | \rvx)$ is a flow (Figure \ref{fig:binary_checkerboard_density_plot_diagonal}), the flexible dequantizer compensates the limitations of the density model by shaping itself to the limitations of the simple distribution. A variant where the density model $p(\rvv)$ is slightly more flexible can be seen when the dequantizer $p(\rvv)$ is a Gaussian with covariance, and $q(\rvv | \rvx)$ is still a flow (Figure \ref{fig:binary_checkerboard_density_plot_cov}). Aided by the dequantizer, the model $p(\rvv)$ aims to place density on the diagonal line which improves the performance to 1.08 bits, which is already close to the theoretical limit. The best performance is achieved when both $q(\rvv|\rvx)$ and $p(\rvv)$ are flexible (Figure \ref{fig:binary_checkerboard_density_plot_flexible}). For this problem we observe the density contracts somewhat away from boundaries, and the center has relatively high density.

\begin{table}
\caption{Binary checkerboard \textit{vi}-dequantization performance for different dequantizer $q(\rvv|\rvx)$ and density model $p(\rvv)$ pairs in bits. Lower is better.}
    \label{tab:results_binary_checkerboard}
    \centering
    \begin{tabular}{l l l l l}
         \toprule
         & & \multicolumn{3}{c}{$q(\rvv|\rvx)$} \\
         & & Uniform & Diag. & Flow \\ \midrule
        \multirow{3}{*}{\rotatebox[origin=c]{90}{$p(\rvv)$}}
        & Diag. & 2.51 & 2.08 & 2.01 \\
        & Cov. & 1.91 & 1.66 & 1.08 \\
        & Flow & 1.11 & 1.02 & 1.02 \\ \bottomrule
    \end{tabular}
\end{table}

\paragraph{Performance} The effects seen in Figure \ref{fig:binary_checkerboard_density_plot} also translate quantitatively with the likelihood performance of these models (Table \ref{tab:results_binary_checkerboard}). Note that the more flexible the distributions, the better the performance. Another interesting observation is that when $p(\rvv)$ is a flow distribution, a Gaussian $q(\rvv|\rvx)$ and a flow $q(\rvv|\rvx)$ have equal performance. Presumably, the flexibility of $p(\rvv)$ does not require a more complicated dequantizer for this relatively simple problem, where performance is already close to the theoretical limit of 1 bit.

Next to \textit{vi} dequantization, we study the effects when models are optimized using \textit{iw} and \textit{R\'{e}nyi} dequantization (Table \ref{tab:results_iw_renyi_binary_checkerboard}). We find that uniformly dequantized models that are trained using \textit{R\'{e}nyi} dequantization are considerably better than $vi$ in terms of likelihood. Furthermore, when trained using \textit{iw} dequantization the model achieves performance close to the theoretical limit. For more complicated dequantizers though, we find that improvements are negligible. Therefore, these sophisticated objectives appear to be particularly useful when the dequantization distribution is simple. An interesting remark specific to the binary checkerboard, is that we found that \textit{R\'{e}nyi} dequantization for larger values than $M$ = 2 tends to diverge from the \textit{iw} dequantization. The model fits to this divergence which results in a worse likelihood score. Presumably, this occurs because the binary checkerboard is low-dimensional, as this effect is not seen on binary MNIST and CIFAR10 (see the following subsections). Additionally, we note that ARD and bipartite dequantization are equivalent in two dimensions, and hence comparison is only meaningful on higher dimensional problems. 

\begin{table}
\caption{Likelihood performance on binary checkerboard when trained with \textit{iw} or \textit{R\'{e}nyi} dequantization objectives in bits per dimension (bpd). The reported values are a (bounded) approximation of -~$\log P(\rvx)$ using \textit{iw}-dequantization with 256 samples. Lower is better.}
    \label{tab:results_iw_renyi_binary_checkerboard}
    \centering
    \begin{tabular}{l l l l}
         \toprule
         & \multicolumn{3}{c}{$q(\rvv|\rvx)$} \\
         & Uniform & Normal & Flow \\ \midrule
        \textit{vi}$^{\star}$ & 1.05 & 1.00 & 1.00 \\
        \textit{iw} ($M = 16$) & 1.00 & 1.00 & 1.00 \\
        \textit{R\'{e}nyi} ($M = 2$) & 1.02 & 1.00 & 1.01 \\ \bottomrule
    \end{tabular}
    \begin{flushleft}
    $^\star$ \scriptsize{\textit{vi} is equivalent to \textit{iw} or \textit{R\'{e}nyi} with $M = 1$.}
    \end{flushleft}
\end{table}

\subsection{Image distribution modelling}
In this section different dequantizer distributions and objectives are tested on binary MNIST and CIFAR10 (8 bit and 5 bit). Problems with lower bit depths are interesting, because dequantization noise is relatively large with respect to the range of values that the data takes. 

\paragraph{Importance weighted and Renyi dequantization}
Similar to the 2d example, more sophisticated training objectives are most advantageous when dequantization distributions are simple, which can be seen in Table \ref{tab:iw_renyi_bmnist_cifar}. On binary MNIST, training using \textit{iw} dequantization improves negative likelihood performance from 0.162 bpd to 0.159 bpd. Again, for more expressive learnable dequantizers, we find that the added benefit of these objectives is negligible. For 5 and 8 bit CIFAR10 we train only last 100 epochs with the sophisticated objectives and the first with \textit{vi} to reduce the computational cost. We find that for CIFAR10 the performance gains are minimal, possibly due to the higher bit depth. For simple dequantizers on data with low bit depth, $iw$-dequantization considerably improves performance. \textit{R\'{e}nyi} dequantization achieves similar but slightly worse performance, which is acceptable since it is a faster approximation.

\begin{table}
\caption{Likelihood performance for models trained with \textit{iw} or \textit{R\'{e}nyi} objectives and uniform dequantization on binary MNIST and CIFAR10 in bits per dimension (bpd). The reported values are a (bounded) approximations of -~$\log P(\rvx)$ using \textit{iw}-dequantization with 256 samples. Lower is better.}
    \label{tab:iw_renyi_bmnist_cifar}
    \centering
    \scalebox{0.9}{
    \begin{tabular}{l c c c}
         \toprule
         Dataset & bMNIST & \multicolumn{2}{c}{CIFAR10} \\
         bit depth & 1 bit & 5 bit & 8 bit \\\midrule
        \textit{vi}$^{\star}$ & 0.162 & 1.61 & 3.26 \\
        \textit{iw} ($M = 4$) & \textbf{0.159} & 1.61 & 3.25 \\
        \textit{R\'{e}nyi} ($M = 4$) & 0.160 & 1.61 & 3.25 \\ \bottomrule
    \end{tabular}}
    \begin{flushleft}
    $^\star$ \scriptsize{\textit{vi} is equivalent to \textit{iw} or \textit{R\'{e}nyi} with $M = 1$.}
    \end{flushleft}
\end{table}

\begin{table}[t]
\caption{Performance of \textit{vi} dequantization on binary MNIST for different density model $p(\rvv)$ and dequantizer distributions $q(\rvv|\rvx)$ pairs. -~$\log P(\rvx)$ is approximated using 256 importance weighted samples. $\mathrm{KL}(q_{\phi}|p_{\theta})$ is the difference between -~$\log P(\rvx)$ and \textit{vi}. In bits per dimension, lower is better.}
    \label{tab:bmnist_vi_with_likelihood}
    \centering
    \scalebox{0.85}{
    \begin{tabular}{l l l l l l l}
         \toprule
        & & & \multicolumn{4}{c}{$q(\rvv|\rvx)$} \\
        & & &  Uniform & Normal & Bipartite & \textit{ARD} \\ \midrule
       \multirow{6}{*}{\rotatebox[origin=c]{90}{$p(\rvv)$}}
       & \multirow{3}{*}{\rotatebox[origin=c]{90}{Cov.}} &    $\mathrm{KL}(q_{\phi}|p_{\theta})$ & 0.061 & 0.046 & 0.010 & \textbf{0.007} \\
        & & $vi$ & 0.533 & 0.268 & 0.196 & \textbf{0.190} \\
        & & $-\log P(x)$ & 0.472 & 0.242 & 0.186 & \textbf{0.183} \\
        \cmidrule{3-7}
        & \multirow{3}{*}{\rotatebox[origin=c]{90}{Flow}} & $\mathrm{KL}(q_{\phi}|p_{\theta})$ & 0.014 &                   0.007 & \textbf{0.005} & \textbf{0.005} \\
        & & $vi$ & 0.176 & 0.156 & 0.153 & \textbf{0.152} \\
        & & $-\log P(x)$ & 0.162 & 0.149 & 0.148 & \textbf{0.147} \\ \bottomrule
    \end{tabular}
    }
\end{table}

\begin{table}[t]
\caption{Performance of \textit{vi} dequantization on CIFAR10, in 8 and 5 bit for a flow-based density model $p(\rvv)$ is and different dequantizer distributions $q(\rvv|\rvx)$. -~$\log P(\rvx)$ is approximated using 256 importance weighted samples. $\mathrm{KL}(q_{\phi}|p_{\theta})$ is the difference between -~$\log P(\rvx)$ and \textit{vi}. In bits per dimension, lower is better.}
    \label{tab:cifar10_vi_with_likelihood}
    \centering
    \scalebox{0.9}{
    \begin{tabular}{l l l l l l}
         \toprule
        & & \multicolumn{4}{c}{$q(\rvv|\rvx)$} \\
         & &  Uniform & Normal & Bipartite & \textit{ARD} \\ \midrule
        \multirow{3}{*}{\rotatebox[origin=c]{90}{8 bit}}
        & $\mathrm{KL}(q_{\phi}|p_{\theta})$ & 0.03 & \textbf{0.02} & \textbf{0.02} & \textbf{0.02} \\
        & $vi$ & 3.29 & 3.21 & 3.18 & \textbf{3.16} \\
        & $-\log P(x)$ & 3.26 & 3.19 & 3.16 & \textbf{3.14} \\
        \midrule
        \multirow{3}{*}{\rotatebox[origin=c]{90}{5 bit}} & $\mathrm{KL}(q_{\phi}|p_{\theta})$ & 0.04 & 0.02 & \textbf{0.01} & 0.02 \\
        & $vi$ & 1.65 & 1.50 & 1.43 & \textbf{1.41} \\
        & $-\log P(x)$ & 1.61 & 1.48 & 1.42 & \textbf{1.39}\\ \bottomrule
    \end{tabular}
    }
\end{table}

\paragraph{Autoregressive Dequantization}
Experiments show that ARD outperforms all other dequantization distributions, when trained using comparable architectures. On binary MNIST we consider two density models, a Gaussian with covariance and a flow based model. Both these models benefit from ARD as presented in Table \ref{tab:bmnist_vi_with_likelihood}. Similar to findings on the 2D binary checkerboard, when $p(\rvv)$ is a Gaussian and therefore less flexible, more flexible dequantizers can compensate. Consider for instance the performance improvement in log-likelihood from uniform dequantization to ARD: the improvement for a flow is 0.015 bpd, whereas the improvement for the Gaussian is approximately 0.3 bpd.

On CIFAR10, again our proposed ARD outperforms other dequantization methods in likelihood modelling (Table \ref{tab:cifar10_vi_with_likelihood}). Notice that dequantization distribution seems to matter more when bit-depths are smaller. To see this, consider the log-likelihood improvement when comparing uniform dequantization and ARD: For the 8 bit data the improvement is 0.12 bpd, which is about 3.7\% relative to the total bpd. However, for 5 bit data the improvement is already 0.20 bpd which is about 12\% relatively. We observe that log-likelihood modelling of lower bit depth data may especially benefit from more expressive dequantizers.

\paragraph{Literature} In this experiment the model using ARD is compared with other methods in the literature. Experiments show that our model outperforms other methods in the literature on both variational inference objective and negative likelihood (Table \ref{tab:cifar10_literature}). In general we compare to models that do not require an autoregressive inverse to sample from, where the exception is marked \textsuperscript{$\star$}. In particular, we report \textit{vi} evaluation, also referred to as Expected Lower Bound (ELBO), and we report the approximate negative likelihood -~$\log P(\rvx)$ using $1000$ importance weighted samples following \citet{maaloe2019biva}. Note that \citet{ho2019flowpp} use $16384$ samples, which skews the experiment in their favour for \textit{vi} and -~$\log P(\rvx)$, but against them for $\mathrm{KL}(q_{\phi}|p_{\theta})$. Architecturally, the density model in ARD is most similar to IDF \cite{hoogeboom2019integer}, where 1 $\times$ 1 convolutions from Glow \cite{kingma2018glow} and scale transformations from RealNVP \cite{dinh2017density} are added. Flow++ \cite{ho2019flowpp} has additional attention layers and MintNet \cite{song2019mintnet} has autoregressive transformations instead of coupling layers in the density model. Note that even though our model utilizes autoregressive components similar to MintNet \cite{song2019mintnet}, our model is computationally cheap to invert since it does not require the solution to autoregressive inverses. Residual Flow \cite{chen2019residual} utilizes invertible ResNets instead of coupling layers.

\begin{table}
\caption{Comparison of negative log-likelihood, \textit{vi} dequantization (ELBO) evaluation of our model versus literature. -~$\log P(\rvx)$ is approximated using $1000$ importance weighted samples. $\mathrm{KL}(q_{\phi}|p_{\theta})$ is the difference between -~$\log P(\rvx)$ and \textit{vi}. In bits per dimension, lower is better.}
    \label{tab:cifar10_literature}
    \centering
    \scalebox{0.79}{
    \begin{tabular}{p{.275\textwidth} c c c}
         \toprule
        Method & $\mathrm{KL}(q_\phi|p_\theta)$ & $vi$ & $-\log P(x)$ \\ \midrule
        IAF-VAE \citep{kingma2016improving}& 0.04 & 3.15 & 3.11 \\
        BIVA \cite{maaloe2019biva} & 0.04 & 3.12 & 3.08 \\ \midrule
        Glow \citep{kingma2018glow} & \textit{n/a} & 3.35 & \textit{n/a} \\
        FFJORD \citep{grathwohl2019ffjord} & \textit{n/a} & 3.40 & \textit{n/a} \\
        IDF \citep{hoogeboom2019integer} & - & - & 3.32 \\
        MintNet \cite{song2019mintnet}$^{\star}$ & \textit{n/a} & 3.32 & \textit{n/a} \\
        Residual Flow \cite{chen2019residual}$^{\dagger}$ &  \textit{n/a} & 3.28 & \textit{n/a} \\
        Flow++ \citep{ho2019flowpp}$^{\dagger}$ & 0.04 & 3.12 & 3.08 \\
        ARD & \textbf{0.03} & \textbf{3.09} & \textbf{3.06} \\ \bottomrule
    \end{tabular}}
\begin{flushleft}
    $\star$ \scriptsize{Sampling from model requires autoregressive inverse.} \\
    $\dagger$ \scriptsize{Sampling from model requires other iterative procedures.} \\
    \textit{n/a} \scriptsize{not available, this value exists but was not reported in the literature.}
    \end{flushleft}
\end{table}

\paragraph{Recommendations} This section aims to give the reader recommendations on what dequantization methods to use and what gains are to be expected. When dequantization noise is fixed, \textit{iw} or \textit{R\'{e}nyi} dequantization objectives generally improve log-likelihood performance. When these objectives are too expensive to utilize for the complete procedure, it is generally enough to train the first epochs/iterations with \textit{vi} dequantization, and then finetune the last epochs using \textit{iw} or \textit{R\'{e}nyi}. Note that by design of the objectives, the approximate posterior $q(\rvv | \rvx)$ will diverge more from the (unknown) true posterior $p(\rvv | \rvx)$. Therefore, a downside of these objectives is that a single sample \textit{iw} dequantization (equivalent to \textit{vi}) will be a poor approximation to the log-likelihood, and instead multiple samples are required. 

When dequantization noise can be learned, the \textit{vi} dequantization objective is generally sufficient. If the reader either has a simple density model or is interested in obtaining the highest log-likelihood performance possible, we recommend using ARD, as its flexibility improves the modelling performance. However when computational resources are scarce and some performance decrease is acceptable, Gaussian dequantization might be a good simple alternative.

\section{Conclusion}

In this paper we propose two dequantization objectives: importance-weighted (\textit{iw}) dequantization and R\'{e}nyi dequantization. In addition, we improve the flexibility of dequantization distributions with autoregressive dequantization (ARD). Empirically we demonstrate improved likelihood modelling for models trained with \textit{iw} and R\'{e}nyi dequantization when dequantization distributions are simple. Furthermore we demonstrate that ARD achieves a negative log-likelihood of 3.06 bits per dimension on CIFAR10, which to the best of our knowledge is state-of-the-art among distribution models that do not require autoregressive inverses for sampling. 





\newpage
\bibliography{references}
\bibliographystyle{icml2020}

\clearpage

\appendix

\section{Architecture and Optimization details}
\label{sec:architecture_and_optimization_details}

\begin{table*}
\caption{Optimization details.}
    \label{tab:optimization_details}
    \centering
    \scalebox{0.85}{
    \begin{tabular}{l l l l l l l l l}
         \toprule
         Experiment & levels & subflows & net. depth & net. channels & context channels & $q$ levels & $q$ subflows & batch size \\ \midrule
        Binary checkerboard & 1 & 8 & 12 & 192 & 16 & 1 & 4 & 128\\
        Binary MNIST & 2 & 8 & 12 & 192 & 16 & 1 & 4 & 128\\
        CIFAR10 5bit & 2 & 10 & 12 & 768 & 16 & 1 & 2 & 256\\ 
        CIFAR10 & 2 & 10 & 12 & 768 & 16 & 1 & 2 & 256 \\ 
        CIFAR10 (Literature comparison) & 2 & 18 & 12 & 768 & 16 & 1 & 2 & 128 \\ \bottomrule
    \end{tabular}}
\end{table*}

Models were all optimized using \cite{kingma2015adam} with a learning rate of $0.0005$ and standard $\beta$ parameters. Furthermore, during initial 10 epochs the learning rate is multiplied by epoch divided by 10, referred to as \textit{warmup} \cite{kingma2018glow}. All our code was implemented in PyTorch \cite{paszke2017automatic}. The basic architecture was built following \cite{kingma2018glow}: The flow is divided in multiple levels with a decreasing number of dimensions. At the end of every level, half of the representation is modelled using a factor out layer (splitprior) \cite{dinh2017density,kingma2018glow}. Every level consists of subflows, \emph{i.e.} a coupling layer followed by a 1 $\times$ 1 convolution \cite{kingma2018glow}. The coupling layers utilize neural networks as described in \cite{hoogeboom2019integer}. For the autoregressive transformation, we utilize the masking as described in \cite{song2019mintnet}. In terms of autoregressive order, this is equivalent to reshaping a C $\times$ H $\times$ W image to a vector and applying the autoregressive mask. This is opposed to masking in \cite{kingma2016improving}, which is equivalent to a mask on a reshaped H $\times$ W $\times$ C image. In practice, the autoregressive transformation is obtained by masking convolutions.

\section{Samples from trained model}
\begin{figure}
    \centering
    \includegraphics[width=0.4\textwidth]{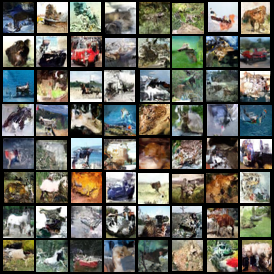}
    \caption{Samples from the flow model in the literature comparison, trained using \textit{ARD}.}
    \label{fig:samples_cifar10}
\end{figure}

Visualization of samples $\rvv \sim p(\rvv)$ from a density model, and the quantizer $\rvx \sim P(\rvx | \rvv)$ are depicted in Figure \ref{fig:samples_cifar10}. The quantizer is simply a Kronecker delta peak and amounts to applying a floor function. The density model is a flow trained with autoregressive dequantization on standard 8 bit CIFAR10. Notice that although the method is trained using autoregressive dequantization, the density model $p(\rvv)$ uses bipartite transformations and does not require the solution to autoregressive inverses.

\section{Visualizations on Binary Checkerboard}
\label{sec:visualizations}
In this section a comprehensive overview of the distributions dequantizer and density model pairs is visualized. The models trained using variational inference are displayed in Table \ref{tab:visualizations}. In general, the dequantizer $q(\rvv|\rvx)$ and density model $p(\rvv)$ try to compensate for each other where they are lacking flexibility. This effect can be seen when $q(\rvv|\rvx)$ is a flow and $p(\rvv)$ is a diagonal Gaussian, a covariance Gaussian and lastly a flow. When $p(\rvv)$ is a flow, it is generally difficult to capture the edges of the squares when dequantization noise is uniform. However, both Gaussian and flow dequantization perform equally when the model $p(\rvv)$ is a flow. In this simple problem, Gaussian dequantization is sufficiently flexible when combined with a flow. 

The models trained using \textit{iw} and \textit{R\'{e}nyi} dequantization objectives are depicted in Table \ref{tab:visualizations_iw_renyi}. An important difference with \textit{vi}-dequantization is that it is much less important for $q(\rvv|\rvx)$ and $p(\rvv)$ to match completely. Rather, more emphasis is placed so that $p(\rvv)$ places distribution somewhere in the appropriate bin, where the exact location in the bin matters less. As a result, when $q(\rvv|\rvx)$ is uniform the model $p(\rvv)$ is not forced to learn the uniform square and retracts somewhat away from the edges. 

\begin{table*}[b]
\caption{Different dequantizer $q(\rvv|\rvx)$ and density model $p(\rvv)$ pairs trained using \textit{vi}-dequantization. The depicted values are computed using $vi$-dequantization (ELBO).}
    \label{tab:visualizations}
    \centering
    \begin{tabular}{c c c c c}
         \toprule
         & & \multicolumn{3}{c}{$q(\rvv|\rvx)$} \\
         & & Uniform & Normal & Flow \\ \midrule
        & Normal diag. & 
        \makecell{
        \includegraphics[align=c, width=0.24\textwidth]{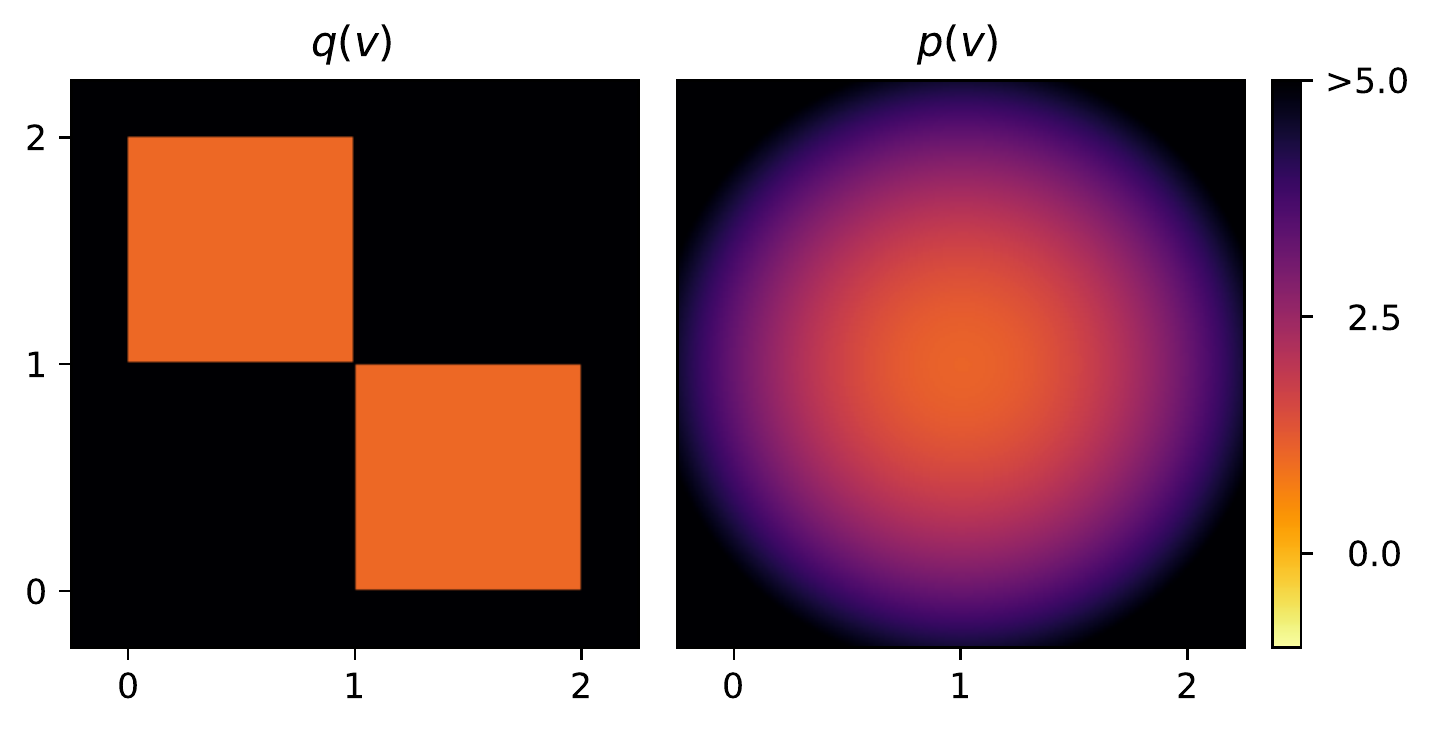}
        \\ 2.51}
        &
        \makecell{\includegraphics[align=c, width=0.24\textwidth]{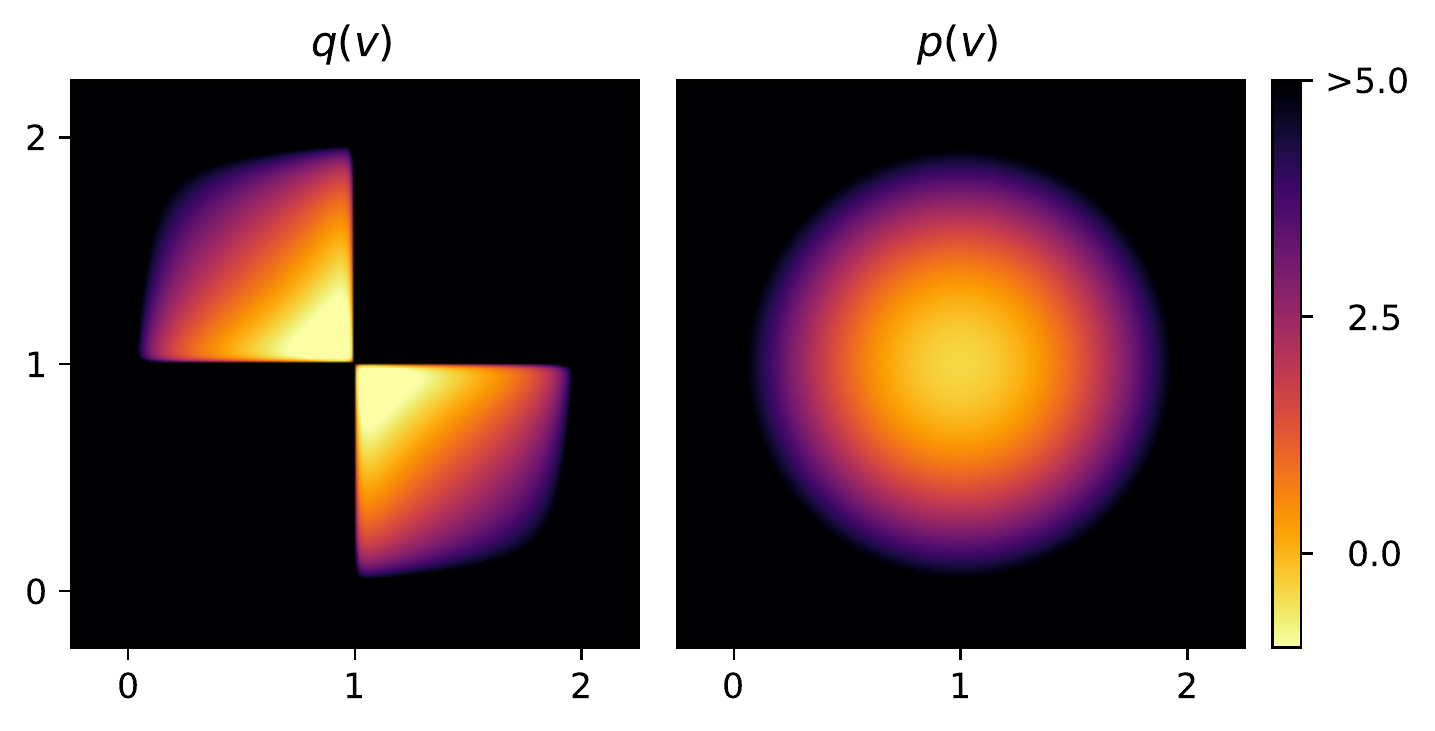} \\ 2.08}
        &
        \makecell{\includegraphics[align=c, width=0.24\textwidth]{images/px_diagonalgaussian_qu_x_bi_affine_normalize_linear_vi_1.pdf}
        \\ 2.01} \\
        {\rotatebox[origin=b]{90}{$p(\rvv)$}} & Normal cov. &
        \makecell{\includegraphics[align=c, width=0.24\textwidth]{images/px_covariancegaussian_qu_x_uniform_normalize_linear_vi_1.pdf}
        \\ 1.91}
        &
        \makecell{\includegraphics[align=c, width=0.24\textwidth]{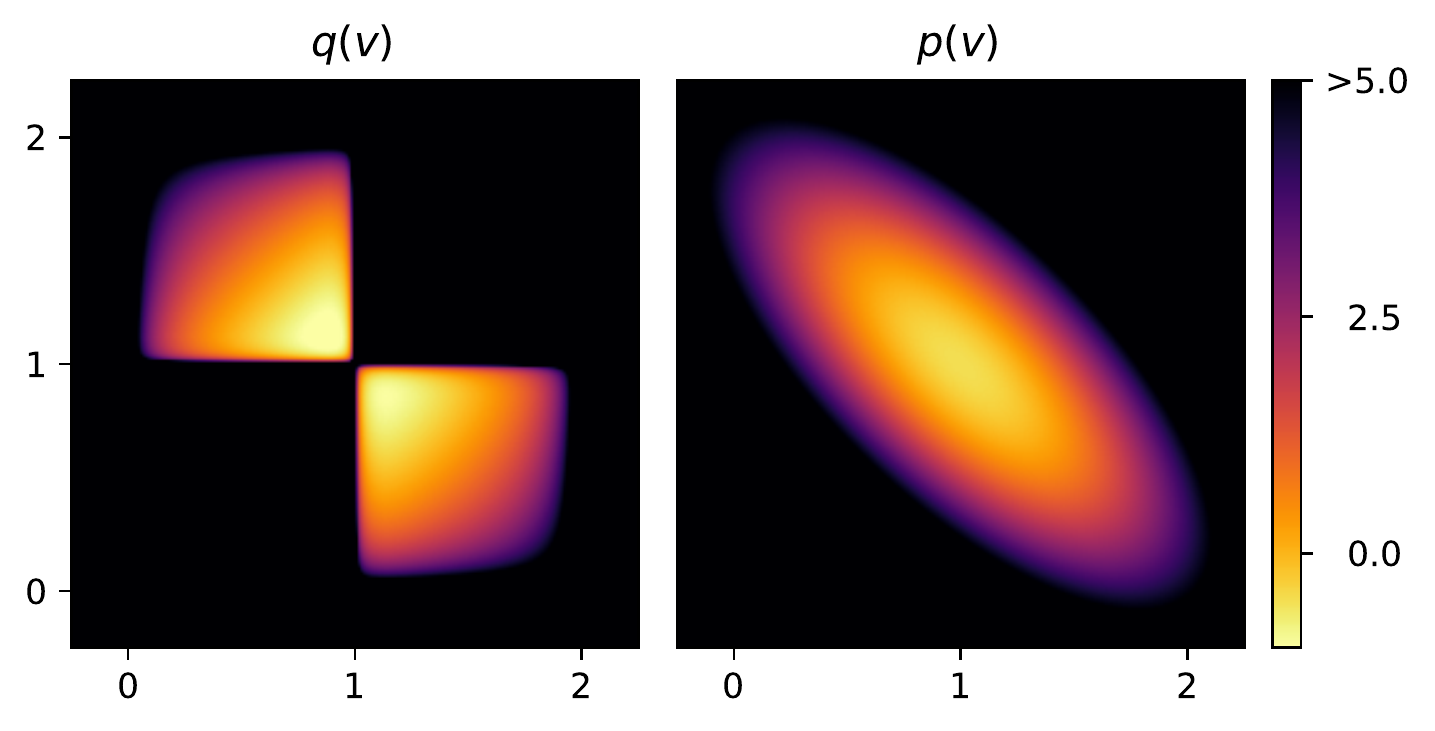}
        \\1.66} &
        \makecell{\includegraphics[align=c, width=0.24\textwidth]{images/px_covariancegaussian_qu_x_bi_affine_normalize_linear_vi_1.pdf}
        \\ 1.08} \\
        & Flow &
        \makecell{\includegraphics[align=c, width=0.24\textwidth]{images/px_bi_affine_qu_x_uniform_normalize_linear_vi_1.pdf}
        \\ 1.11} &
        \makecell{\includegraphics[align=c, width=0.24\textwidth]{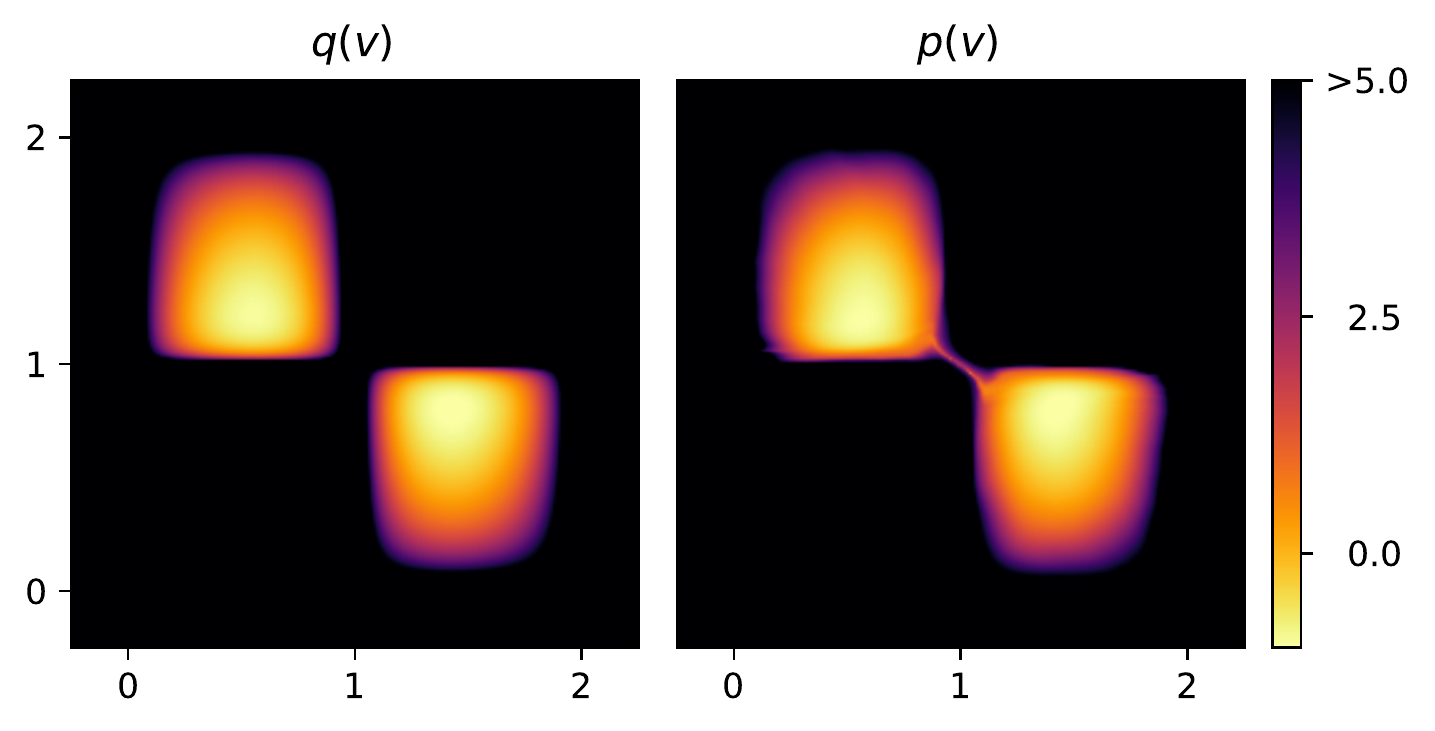}
        \\ 1.02} &
        \makecell{\includegraphics[align=c, width=0.24\textwidth]{images/px_bi_affine_qu_x_bi_affine_normalize_linear_vi_1.pdf}
        \\ 1.02} \\ \bottomrule
    \end{tabular}
\end{table*}

\begin{table*}[t]
\caption{Models trained using \textit{iw} and \textit{R\'{e}nyi} dequantization with different dequantizing distributions $q(\rvv|\rvx)$, and a flow $p(\rvv)$. The values are an approximation of - $\log P(x)$ using importance-weighted dequantization with $M_{\mathrm{test}} = 256$ samples.}
    \label{tab:visualizations_iw_renyi}
    \centering
    \begin{tabular}{c c c c}
         \toprule
         & \multicolumn{3}{c}{$q(\rvv|\rvx)$} \\
         & Uniform & Normal & Flow \\ \midrule
        \textit{vi} &
        \makecell{\includegraphics[align=c, width=0.24\textwidth]{images/px_bi_affine_qu_x_uniform_normalize_linear_vi_1.pdf}
        \\ 1.05} &
        \makecell{\includegraphics[align=c, width=0.24\textwidth]{images/px_bi_affine_qu_x_diagonalgaussian_normalize_linear_vi_1.pdf}
        \\ 1.00} &
        \makecell{\includegraphics[align=c, width=0.24\textwidth]{images/px_bi_affine_qu_x_bi_affine_normalize_linear_vi_1.pdf}
        \\ 1.00} \\
        \textit{iw} ($M=16$) & 
        \makecell{\includegraphics[align=c, width=0.24\textwidth]{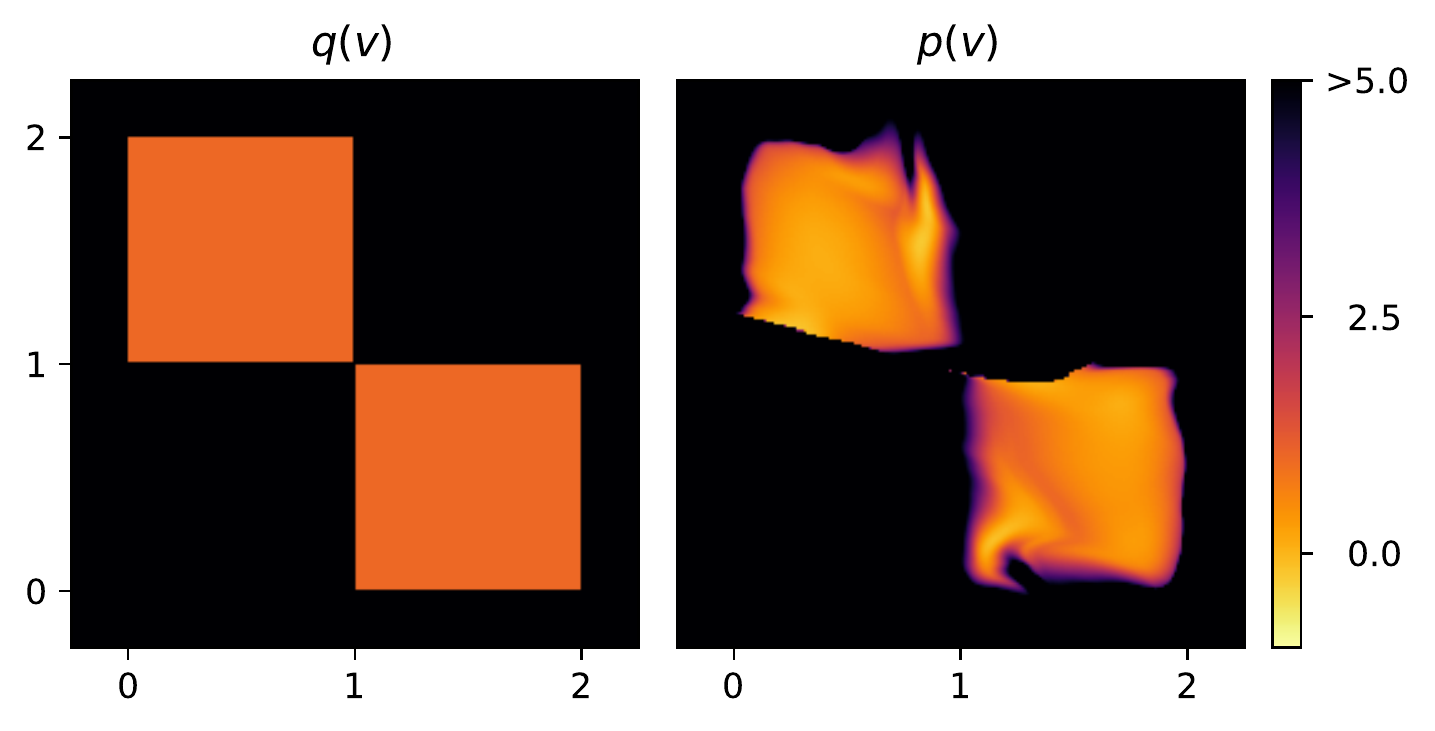}
        \\ 1.00}
        &
        \makecell{\includegraphics[align=c, width=0.24\textwidth]{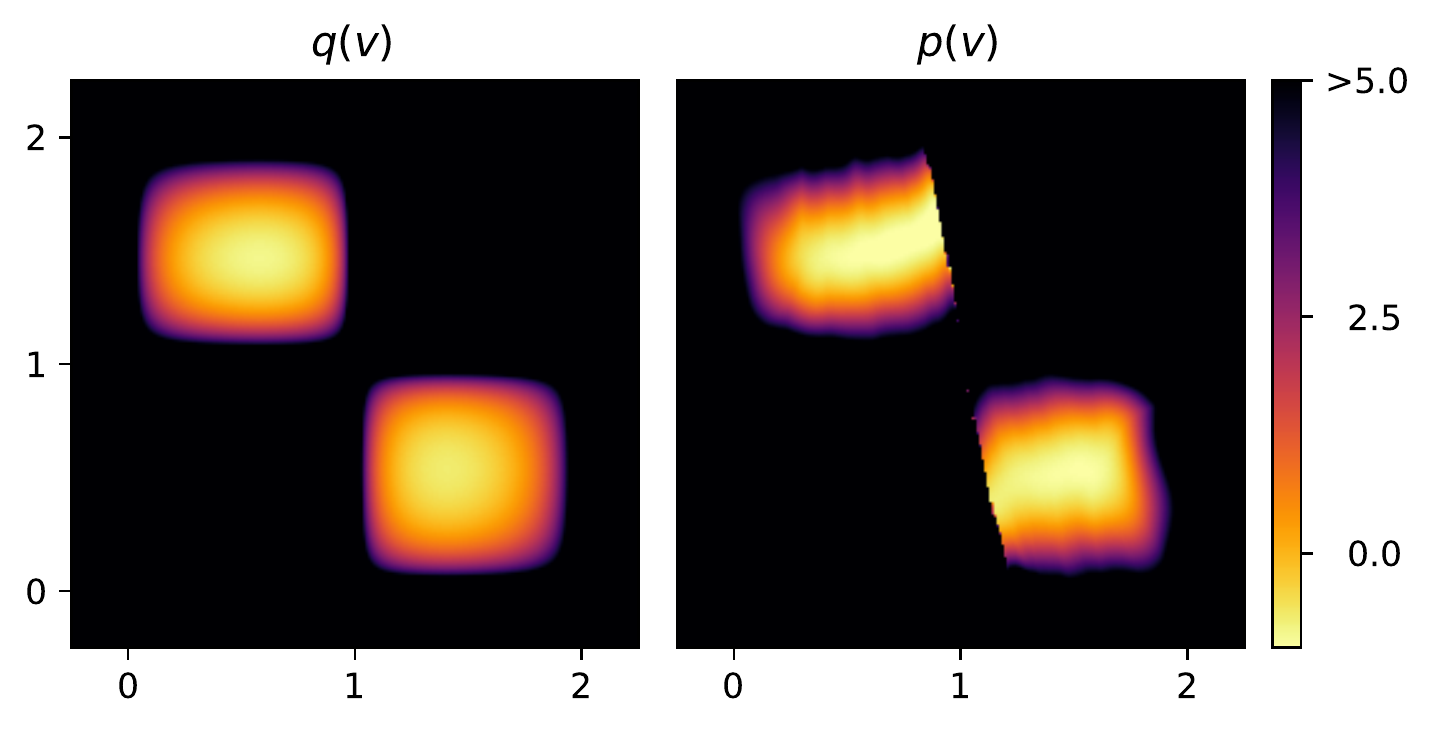}
        \\ 1.00} &
        \makecell{\includegraphics[align=c, width=0.24\textwidth]{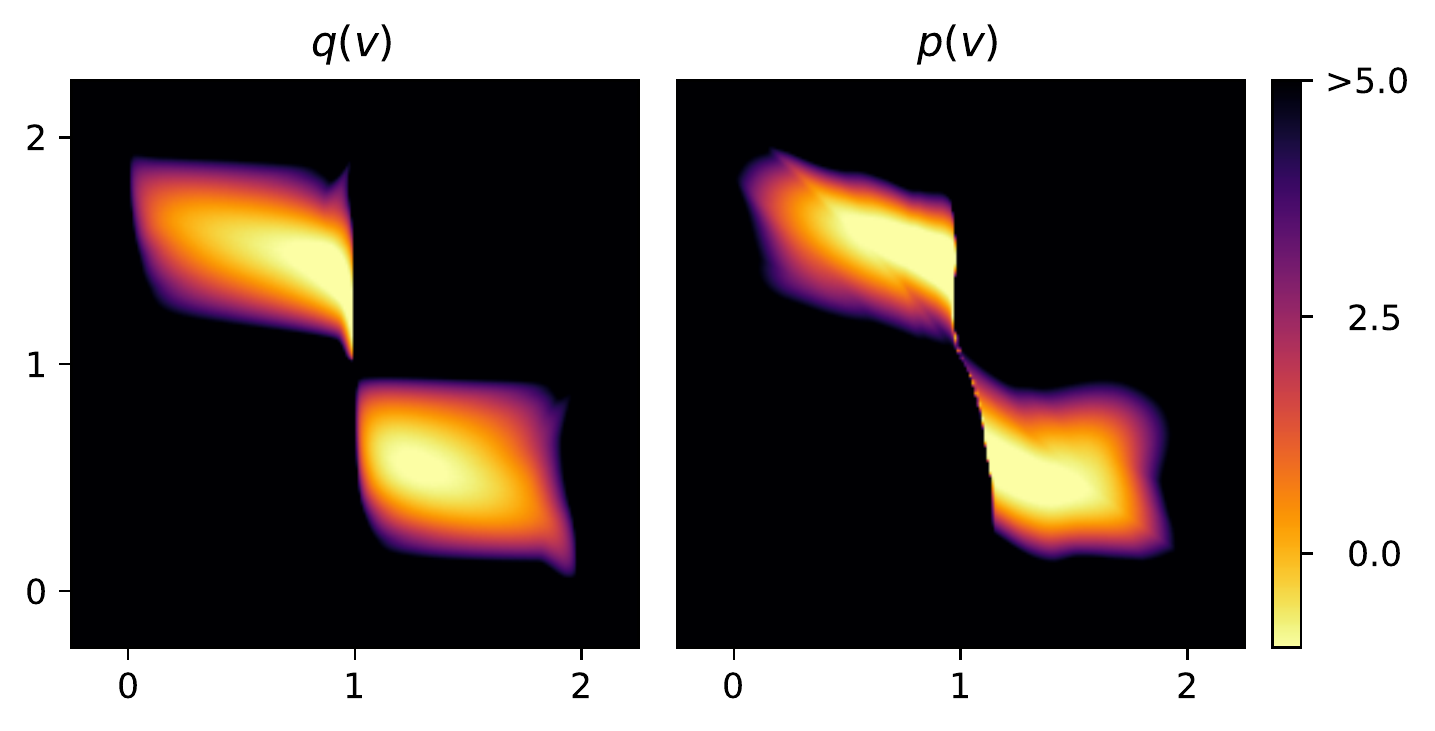}
        \\ 1.00} \\
        \textit{R\'{e}nyi} ($M=2$) &
        \makecell{\includegraphics[align=c, width=0.24\textwidth]{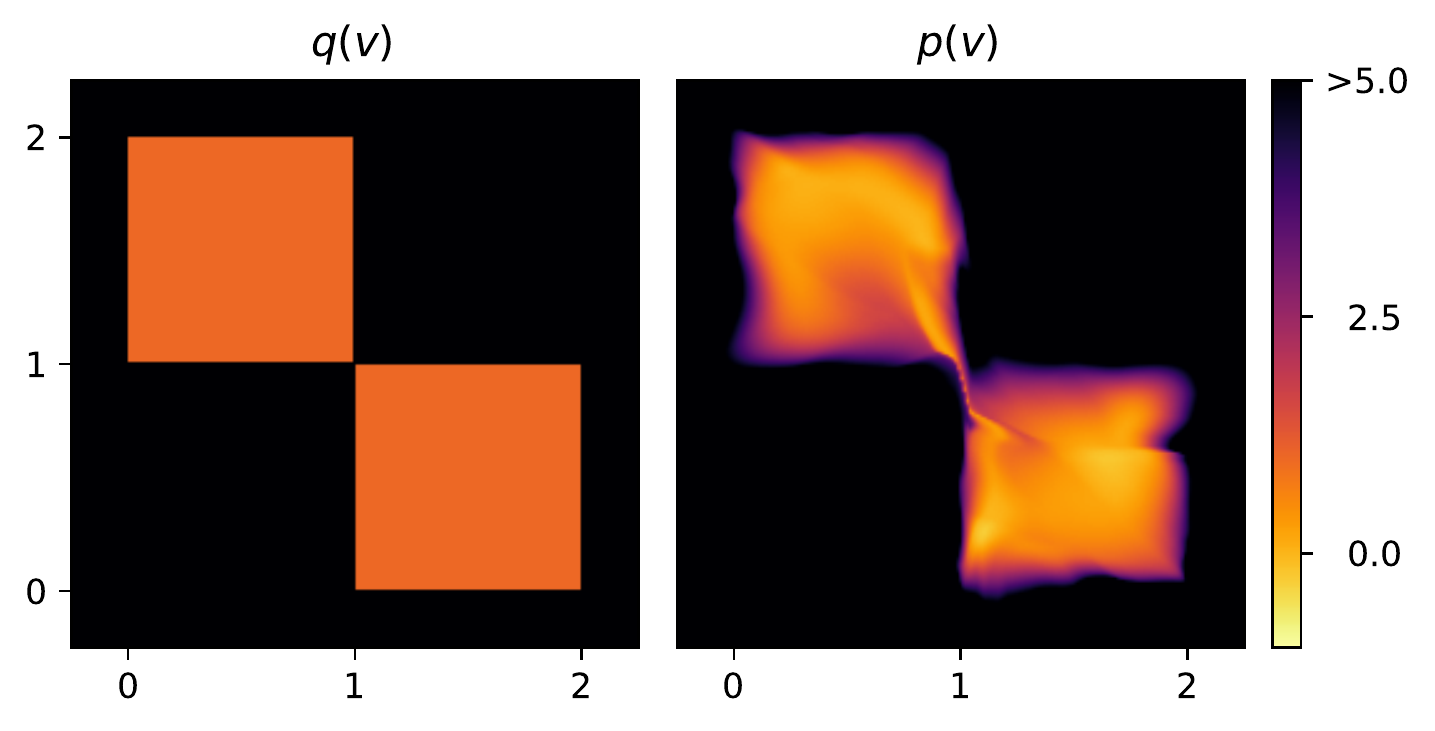}
        \\ 1.02} &
        \makecell{\includegraphics[align=c, width=0.24\textwidth]{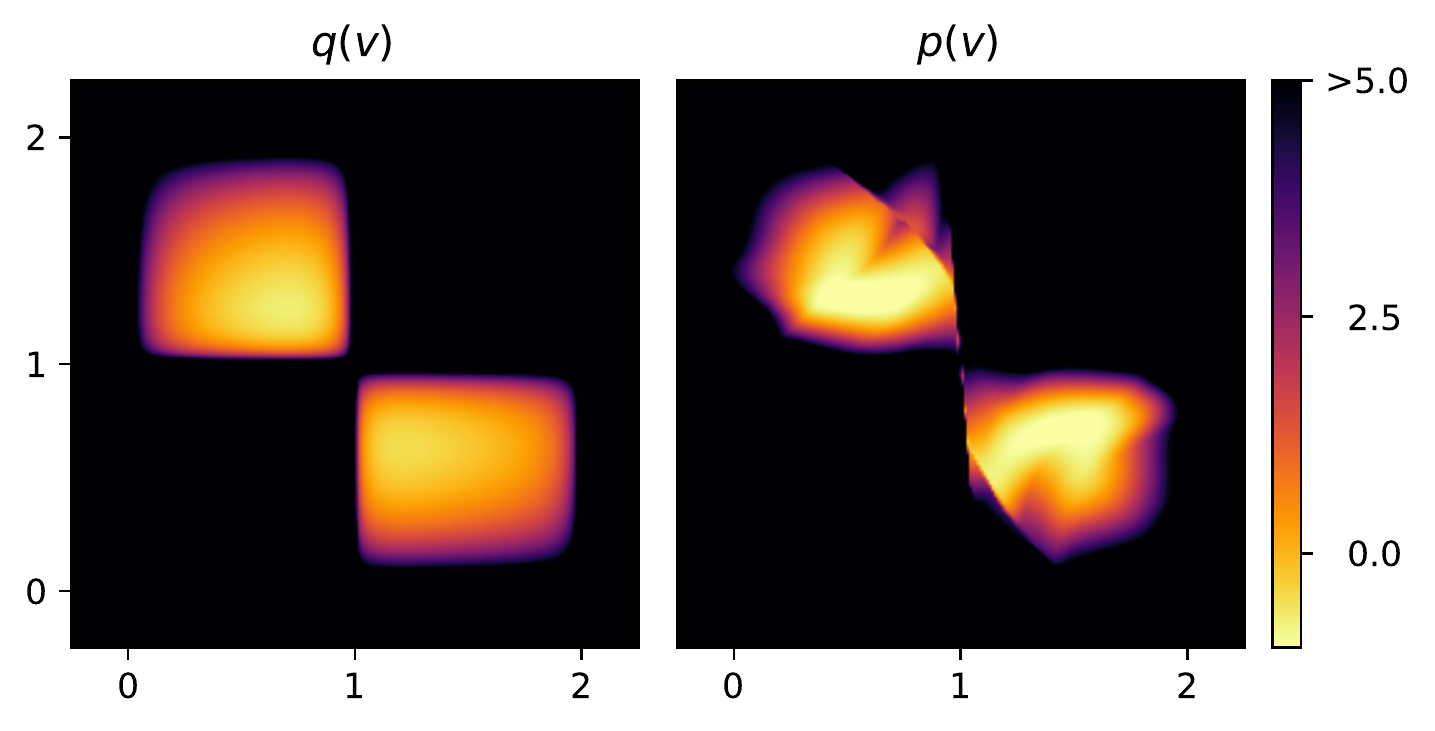}
        \\ 1.00} &
        \makecell{\includegraphics[align=c, width=0.24\textwidth]{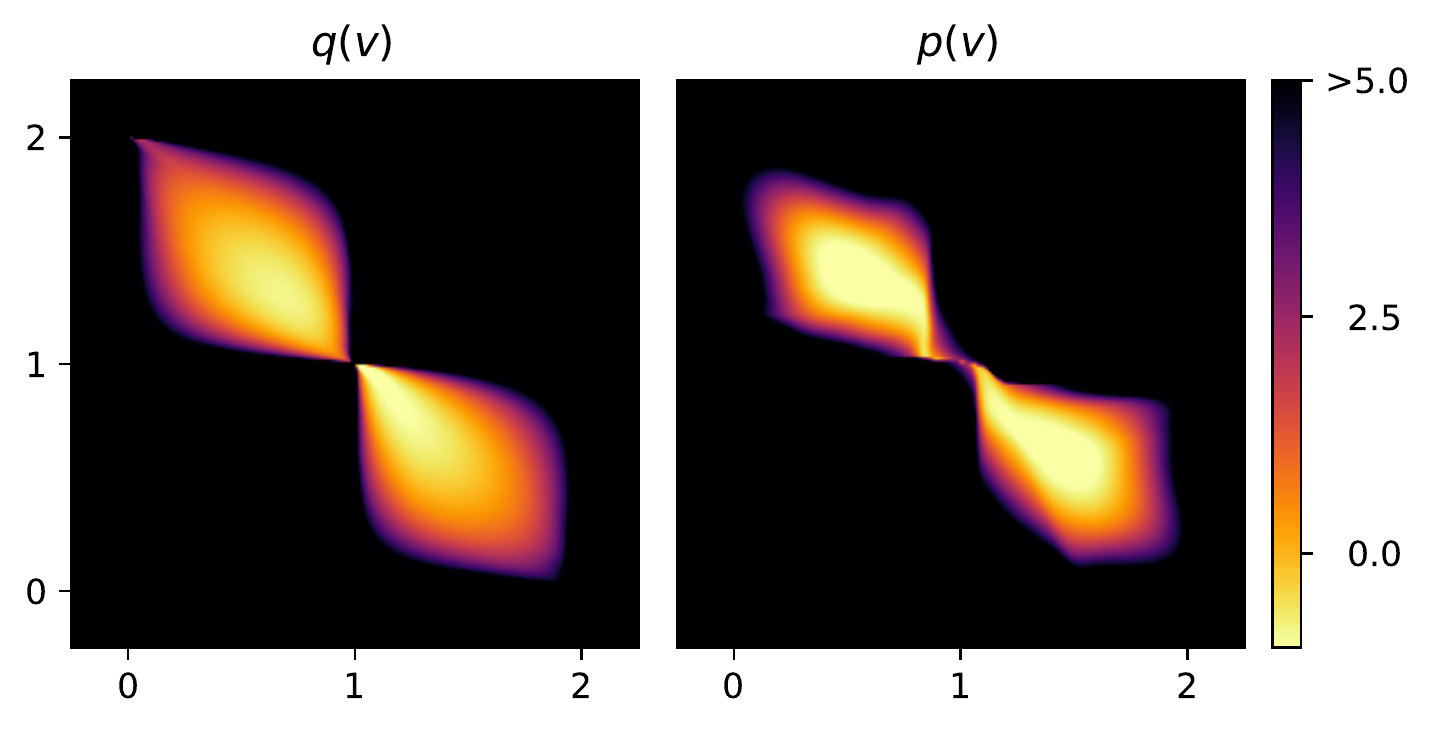}
        \\ 1.01} \\
        \bottomrule
    \end{tabular}
\end{table*}





\end{document}